%% file: main.tex
\pdfoutput=1

\documentclass[11pt]{article}

\usepackage[final]{acl}

\usepackage{times}
\usepackage{latexsym}
\usepackage{hyperref}
\usepackage{url}

\usepackage[T1]{fontenc}

\usepackage[utf8]{inputenc}

\usepackage{microtype}

\usepackage{inconsolata}

\usepackage{subcaption}
\usepackage{amsmath}
\usepackage{amsfonts}
\usepackage{tcolorbox}
\usepackage{booktabs}
\usepackage{algorithm}
\usepackage{multirow}
\usepackage[noend]{algpseudocode}
\usepackage{amsmath, amsthm, amssymb, amsfonts} 
\usepackage{comment}
\usepackage{wrapfig}
\newcommand{\change}[1]{{#1}}
\newcommand{\finalchange}[1]{{#1}}

\title{LLM Performance Predictors are good initializers for Architecture Search}

\author{\textbf{Ganesh Jawahar}$^{\mu\heartsuit}$\thanks{Work was done while Ganesh was at UBC.} \quad \textbf{Muhammad Abdul-Mageed}$^{\mu\diamondsuit}$ \\ \textbf{Laks V. S. Lakshmanan}$^\mu$ \quad \textbf{Dujian Ding}$^\mu$\\
 $^\mu$University of British Columbia \quad $^\diamondsuit$MBZUAI \quad $^\heartsuit$Google DeepMind \\
\texttt{ganeshjwhr@gmail.com, muhammad.mageed@ubc.ca, \{laks,dujian\}@cs.ubc.ca }}

\begin{document}
\maketitle

\input{tex/abstract}
\input{tex/intro}
\input{tex/related_work}
\input{tex/problem}

\input{tex/supernet_estimator}
\input{tex/gpt_estimator}

\input{tex/search_initializer}

\input{tex/conclusion}
\input{tex/limitations}

\bibliography{anthology,custom}

\appendix

\input{tex/appendix}

\end{document}

%% file: tex/abstract.tex
\begin{abstract}

In this work, we utilize Large Language Models (LLMs) for a novel use case: constructing Performance Predictors (PP) that estimate the performance of specific deep neural network architectures on downstream tasks. We create \textit{PP prompts} for LLMs, comprising (i) \textit{role} descriptions, (ii) \textit{instructions} for the LLM, (iii) \textit{hyperparameter} definitions, and (iv) \textit{demonstrations} presenting sample architectures with efficiency metrics and `training from scratch' performance. In machine translation (MT) tasks, GPT-4 with our PP prompts (LLM-PP) achieves a SoTA mean absolute error and a slight degradation in rank correlation coefficient compared to baseline predictors. Additionally, we demonstrate that predictions from LLM-PP can be distilled to a compact regression model (LLM-Distill-PP), which surprisingly retains much of the performance of LLM-PP. This presents a cost-effective alternative for resource-intensive performance estimation. Specifically, for Neural Architecture Search (NAS), we introduce a \textit{Hybrid-Search} algorithm (HS-NAS) employing LLM-Distill-PP for the initial search stages and reverting to the baseline predictor later. HS-NAS performs similarly to SoTA NAS, reducing search hours by approximately 50\%, and in some cases, improving latency, GFLOPs, and model size. \finalchange{The code can be found at: \url{https://github.com/UBC-NLP/llmas}.}

\end{abstract}

%% file: tex/intro.tex
\section{Introduction}
\label{sec:intro}

Large language models (LLMs) have diverse applications, encompassing both open-ended tasks (e.g., brainstorming and chat) and closed-ended tasks (e.g., summarization and question answering). This study explores a unique application of LLMs: constructing a performance predictor (LLM-PP) for a deep neural network (DNN) architecture. The predictor takes the DNN architecture description, typically hyperparameters (e.g., \#layers, \#attention heads), as input and predicts the performance (e.g., BLEU score) for a specific downstream task. The aim is to create a performance predictor with low prediction errors compared to training from scratch. The hypothesis is that LLMs possess a `general understanding' of DNN architectures, derived from relevant training data like DNN research papers and GitHub repositories. The main objective of this work is to leverage this understanding to design accurate, efficient, and broadly applicable performance predictors, beneficial for tasks like neural architecture search (NAS).

\textbf{How to design an accurate performance predictor (PP)?} 
To answer this, we create \textit{PP prompts} precisely specifying the task. These prompts include: (i) \textit{role}: high-level description of the assigned LLM role, (ii) \textit{instructions}: detailed task instructions (e.g., downstream task, architecture, performance/efficiency metric) for the LLM to follow, (iii) \textit{hyperparameters}: definitions of architecture-specific hyperparameters, and (iv) \textit{demonstrations}: supervised examples for the PP task with architecture descriptions and performance metrics (e.g., BLEU score). Using GPT-4~\citep{gpt4} as our primary LLM and WMT datasets for machine translation (MT) tasks, we find that GPT-4 with our PP prompts (LLM-PP) predicts architecture performance with a mean absolute error achieving the state-of-the-art (SoTA) and a slightly lower rank correlation coefficient compared to previous SoTA weight-sharing supernet-based performance predictors~\citep{hat,jawahar2023mixtureofsupernets}.


Using GPT-4 for LLM-PP entails utilizing the GPT-4 API to score each architecture, rendering LLM-PP prohibitively expensive for various use cases. One example is NAS, where PP evaluates approximately 3,000 candidate architectures for each constraint (e.g., latency $\leq$ 100ms)~\citep{hat}. As of August 2023, GPT-4 pricing is 0.03\$ per $1K$ tokens~\footnote{\url{https://openai.com/pricing}}. Assuming PP prompts consume about one-third of $1K$ tokens, the estimated cost is approximately $\sim$30\$ for a single constraint on the target hardware. With varying constraint values (e.g., 100ms, 200ms), constraint types (e.g., latency, FLOPs, memory), and target hardware (e.g., Nvidia A100, Raspberry Pi), the cumulative cost can quickly become exorbitant (e.g., 1,800\$).

\textbf{How to design cost-effective PP?} To answer this, 
we distill LLM-PP performance predictions into a tiny MLP model (LLM-Distill-PP) using architecture descriptions (e.g., hyperparameter lists) as input features. Surprisingly, LLM-Distill-PP can significantly maintain the performance of LLM-PP. Assuming LLM-Distill-PP needs only 3,000 examples, the estimated cost is approximately $\sim$30\$ for a single downstream task, amortized across various constraint values, types, and target hardware.


\textbf{Can LLM-Distill-PP speed up architecture search 
while preserving the efficiency and the quality of SoTA NAS?} To answer this, 
we apply using LLM-Distill-PP as the PP to design efficient MT architectures via SoTA NAS methods like HAT~\citep{hat}. 
We introduce the \textit{Hybrid-Search} algorithm (HS-NAS), where LLM-Distill-PP serves as the PP for the first 15 search iterations, and a weight-sharing supernet (SoTA performance predictor) takes over for the remaining 15 iterations. HS-NAS achieves roughly 50\% faster search than SoTA NAS, maintaining or improving on the performance of architectures designed by SoTA NAS. In some cases, it also yields reduced latency ($\sim$2\%), FLOPs ($\sim$1\%), and model size ($\sim$2\%).

Our main contributions are as follows: 
\begin{enumerate}
\item We propose LLM-PP, leveraging few-shot prompting of LLM for accurate performance predictors, achieving SoTA mean absolute error. 
\item We introduce LLM-Distill-PP, with a better amortized cost than LLM-PP, suitable for PP-heavy use cases. 
\item HS-NAS, a search algorithm, reduces NAS search time by half compared to SoTA, identifying more efficient architectures by leveraging LLM-Distill-PP and SoTA performance estimators. 
\item We provide prompts, training and evaluation data for LLM-Distill-PP models, and code with detailed reproducibility instructions.
\end{enumerate}

%% file: tex/related_work.tex
\section{Related Work}
\label{sec:related_work}

\noindent\textbf{Performance Predictors.} 
In NLP, a common approach to construct performance predictors is training a weight-sharing supernet model, jointly training various architectures by sharing weights with the largest model in the search space~\citep{hat,autodistill,jawahar-etal-2023-automoe,jawahar2023mixtureofsupernets}. 
During each training step, an architecture is randomly selected from the search space, and its corresponding weights are extracted from the largest model's weight matrices. These weights are then trained for the target task. Post-training, architecture performance is predicted by extracting the relevant weights and evaluating on the validation set. Key challenges in supernet training include weight co-adaptation~\citep{bender2018understanding, zhao2021few}, capacity bottleneck~\citep{jawahar2023mixtureofsupernets}, and gradient conflict~\citep{gong2021nasvit}.

\noindent\textbf{NAS for NLP.} 
NAS is a general framework for designing efficient NLP architectures meeting user-defined constraints across various dimensions: (i) \textit{architecture family} (encoder-only~\citep{autotinybert,autodistill,nasbert,magic}, decoder-only~\citep{litetransformersearch}, encoder-decoder~\citep{hat,jawahar-etal-2023-automoe,jawahar2023mixtureofsupernets} without limiting to Transformers), (ii) \textit{constraint types} (latency, FLOPs, model size), and (iii) \textit{tasks} (task-agnostic pretraining~\citep{autodistill,litetransformersearch,jawahar2023mixtureofsupernets}, task-specific training~\citep{hat,jawahar-etal-2023-automoe}). The evolutionary search-based algorithm employs a performance predictor to identify high-quality architectures, utilizing real or predicted efficiency metrics to discard those not meeting specified constraints. 

\noindent\textbf{LLMs for NAS.} 
GENIUS~\citep{genius}, a recent search algorithm for image classification, uses LLMs to generate convolution-based architectures. However, it trains these candidates from scratch, incurring high practical costs. Contrasting with our approach, (i) GENIUS uses LLMs to generate architectures, while we use LLMs to predict their performance, (ii) the search cost for our work is upper bounded by SoTA NAS for MT ($\sim 5$ NVIDIA V100 hours), much more efficient than GENIUS ($\sim 960 $ NVIDIA V100 hours), and (iii) we focus on Transformer-based encoder-decoder architectures for machine translation. For more on the synergy between LLMs and AutoML, see~\citet{tornede2023automl}. 

Additional background on related topics such as LLMs and distillation can be found in~\ref{sec:related_work_extended}.

%% file: tex/problem.tex
\section{Performance Prediction Problem}
\label{sec:problem_defn}

\change{Informally, the performance prediction problem entails providing a DNN architecture description (usually hyperparameters like \#layers, \#attention heads) to the predictor, which then outputs the performance (e.g., BLEU score) for a specified downstream task. An ideal predictor should minimize prediction errors 
compared to the performance achieved through training from scratch.}
Formally, let $T$ represent a downstream task, $\mathcal{A}_T$ its search space of architectures, and $\mathcal{Y}_T \subset \mathcal{R}$ the real space of performance scores. Define $\mathcal{D}_T$ as the data distribution over $\mathcal{A}_T\times\mathcal{Y}_T$. The performance predictor is denoted by $f_T: \mathcal{A}_T \to \mathcal{Y}_T$. The labeled test set $\mathcal{L}_T^{test} = \{(\mathbf{a}_i, p_i)\}_{i=1}^{m} \sim (\mathcal{D})^m_T$ comprises architecture, performance pairs drawn i.i.d. from $\mathcal{D}_T$. $p_i$ is the performance obtained by training the architecture $\mathbf{a}_i$ from scratch to convergence on task $T$ (known as `training from scratch' (TFS) performance).

The performance predictor's quality is assessed using two metrics: Mean Absolute Error (MAE) calculates the mean absolute difference between predictions and their corresponding TFS performances, formalized as $\sum_{(\mathbf{a}_i, p_i) \sim (\mathcal{D})_T} \frac{|f_T(\mathbf{a}_i)-p_i|}{|\mathcal{(D})_T|}$. Kendall rank correlation coefficient is another metric that computes the ranking correlation between a set of predictions and their corresponding TFS performances, formalized as Kendall-Tau$([f_T(\mathbf{a}_1),\dots,f_T(\mathbf{a}_m)], [p_1,\dots,p_m])$. Examples for these metrics are discussed in Section~\ref{sec:metrics_examples}.
Recently, \citet{jawahar2023mixtureofsupernets} emphasized the importance of both MAE and Kendall-Tau metrics in evaluating performance predictor quality. For instance, a predictor with a 38\% better MAE and a 12\% worse Kendall-Tau, compared to a base predictor, led NAS to find an architecture with a 4\% BLEU improvement. Conversely, a predictor with a 5\% worse MAE and a 6\% higher Kendall-Tau resulted in a NAS architecture with a 0.1\% BLEU improvement. \change{Hence, better MAE and better Kendall-Tau are positively correlated with higher-quality architecture.}

%% file: tex/supernet_estimator.tex
\section{Baseline Performance Predictors}
\label{sec:baseline_perf_predictors}


\change{In NAS for NLP literature, }the SoTA method for constructing performance predictors ($f_T$) involves training a weight-sharing supernet model on task $T$. \change{Simply put, a weight-sharing supernet model is the largest model in the search space, capable of parameterizing all architectures via weight sharing. The parameters for a specific architecture are obtained by extracting the relevant rows and columns from the supernet model's weight matrix. Typically, the supernet is trained by iteratively sampling an architecture from the search space and training the extracted weights for that architecture.}
Formalizing the supernet's training objective: Denote the training data distribution as $\mathcal{X}_{train}$. Represent the training sample and label as $x$ and $y$, where $x, y \sim \mathcal{X}_{train}$. $a_{rand}$ is a uniformly sampled architecture from the search space $\mathcal{A}_T$. $a_{large}$ and $a_{small}$ denote the largest and smallest architectures in $\mathcal{A}_T$. The subnet with architecture $a$ is denoted by $s_{a}$, parameterized by the supernet model weights $W$. The training objective of the supernet using sandwich sampling~\citep{bignas} is given by
\begin{multline}
\nonumber
    \min_W \mathbb{E}_{x,y \sim \mathcal{X}_{train}}[  \mathbb{E}_{{a_{rand}} \sim \mathcal{A}} [  \mathcal{L}(s_{a_{rand}}(x;W),y)] \\ + 
     \mathcal{L}(s_{a_{large}}(x;W),y) +  \mathcal{L}(s_{a_{small}}(x;W),y)].
\end{multline}
Hardware-aware Transformers~\citep{hat} employs single-path one-shot (SPOS) optimization~\citep{spos_guo20}, focusing on optimizing only $a_{rand}$ at each training step. Mixture-of-Supernets~\citep{jawahar2023mixtureofsupernets} (MoS) utilizes mixture-of-experts (MoE)~\citep{switch} to enhance the supernet's capacity, with the router specializing weights for each architecture. MoS comes in two variants: layer-wise MoS and neuron-wise MoS, differing in the degree of freedom for weight generation. Both variants of MoS employ sandwich sampling for supernet training.

%% file: tex/gpt_estimator.tex
\section{LLM Performance Predictor (LLM-PP)}
\label{sec:llm_perf_estimators}
\begin{figure}[t]
\includegraphics[width=3.0in,height=4.3in]{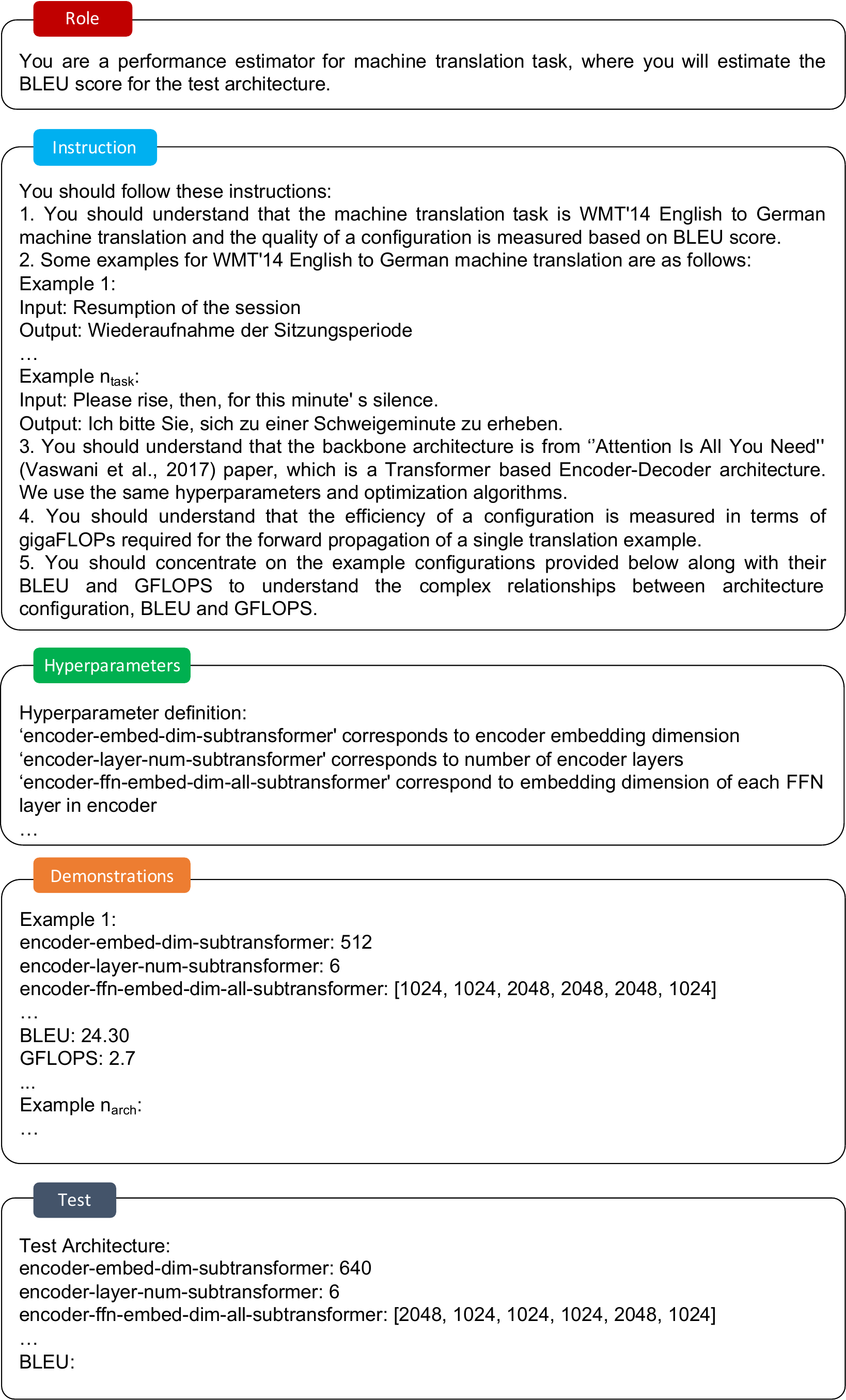}
\caption{Prompt template to prompt LLM to generate performance predictions for WMT'14 EN-DE task. The expanded version of the prompt template can be seen in  Appendix~\ref{sec:prompt_template_expanded}.}
\label{fig:framework_short}
\end{figure}
LLM demonstrates a ``general understanding" of DNN architectures, likely acquired through training on relevant data sources like research papers and GitHub repositories. Testing these architecture understanding capabilities involves prompting LLM to generate hyperparameter definitions and design principles for architecture search~\citep{genius}. These LLM capabilities contribute to effective performance prediction by aiding the mapping of DNN architectures to their performances.

To this end, 
we propose the LLM-based Performance Predictor (LLM-PP), which involves prompting an LLM to generate performance predictions for DNN architectures. The prompts, referred to as \textit{PP prompts}, must be meticulously designed to precisely convey the performance prediction task to the LLM. Illustrated in Figure~\ref{fig:framework_short}, PP prompts break down the task into four main components: {\textcolor{red}{\textit{role}}}, {\textcolor{blue}{\textit{instructions}}}, {\textcolor{green}{\textit{hyperparameters}}}, and {\textcolor{orange}{\textit{demonstrations}}}, followed by the test architecture.

The {\textcolor{red}{\textit{role}}} specifies the LLM's role, describing the downstream task (e.g., machine translation) and the performance metric (e.g., BLEU). The {\textcolor{blue}{\textit{instructions}}} provide five detailed instructions covering the downstream task, DNN architecture, and model efficiency metrics. The first two focus on the task specifics, specifying the task type (e.g., machine translation), dataset (e.g., WMT'14 En-De), performance metric (e.g., BLEU), and inputs/outputs (e.g., source/target language) for $n_{task}$ examples from the dataset. The third instruction details the DNN architecture, including backbone (e.g., Transformer), type (e.g., encoder-decoder), and a reference to the original DNN paper. The fourth instruction outlines efficiency metrics details (e.g., GFLOPs), included in the demonstrations. The final instruction directs the LLM to consider complex relationships between architecture configuration, performance, and efficiency metric. The third component, {\textcolor{green}{\textit{hyperparameters}}}, defines architecture-specific hyperparameters. {\textcolor{orange}{\textit{Demonstrations}}} is the final component containing $n_{arch}$ supervised examples, each representing an architecture from the search space with hyperparameter values, efficiency score, and TFS performance score. The design process of the LLM-PP prompt is discussed in~\ref{sec:prompt_template_design_process}.


\subsection{Evaluation Setup}
\label{sec:llm_pe_eval_setup}
\noindent\textbf{Downstream tasks.} 
We utilize established research~\citep{hat,jawahar-etal-2023-automoe,jawahar2023mixtureofsupernets} and opt for popular machine translation (MT) benchmarks: WMT'14 En-De, WMT'14 En-Fr, and WMT'19 En-De. Detailed statistics of these benchmarks are available in~\ref{sec:mt_dataset_statistics}. Our chosen performance metric is BLEU~\citep{papineni-etal-2002-bleu}.

\noindent\textbf{DNN architecture.} 
We adopt the Transformer-based Encoder-Decoder architecture~\citep{transformer}. The implementation, training settings, and search space ($\mathcal{A}$) mirror~\citet{hat}, detailed in~\ref{sec:mt_train_det_search_space}. Our evaluation dataset (TFS-Eval) is sourced from~\citet{jawahar2023mixtureofsupernets}, featuring 30 architectures with their TFS performance scores for each WMT dataset. FLOPs, latency, and model size computations for architectures are done using the implementation from~\citet{hat}.

\noindent\textbf{Performance predictors.} 
Baseline performance predictors include: (i) HAT~\citep{hat}, (ii) Supernet (Sandwich)~\citep{jawahar2023mixtureofsupernets} (HAT, with sandwich sampling instead of SPOS), (iii) Layer-wise MoS~\citep{jawahar2023mixtureofsupernets}, and (iv) Neuron-wise MoS~\citep{jawahar2023mixtureofsupernets}. We build three LLM-PP variants, utilizing \change{Mistral~\citep{jiang2023mistral} (Mistral-7B-Instruct-v0.1)}, ChatGPT~\citep{chatgpt} (GPT-3.5-turbo, June version), and GPT-4~\citep{gpt4} (June version). For PP prompts, we randomly sample: (i) 5 examples ($n_{task} = 5$) from the downstream task for the second instruction and (ii) 10 examples ($n_{task} = 10$) from TFS-eval for the demonstrations component. The remaining 20 examples from TFS-eval will be used for reporting the predictor quality. For all predictors, we repeat the experiments with three different seeds and report the average MAE and Kendall-Tau between the predictor performance and the TFS performance.

\begin{figure*}[t!]
    \centering
    \begin{subfigure}[t]{0.3\textwidth}
        \centering
        \includegraphics[height=1.3in, width=1.9in]{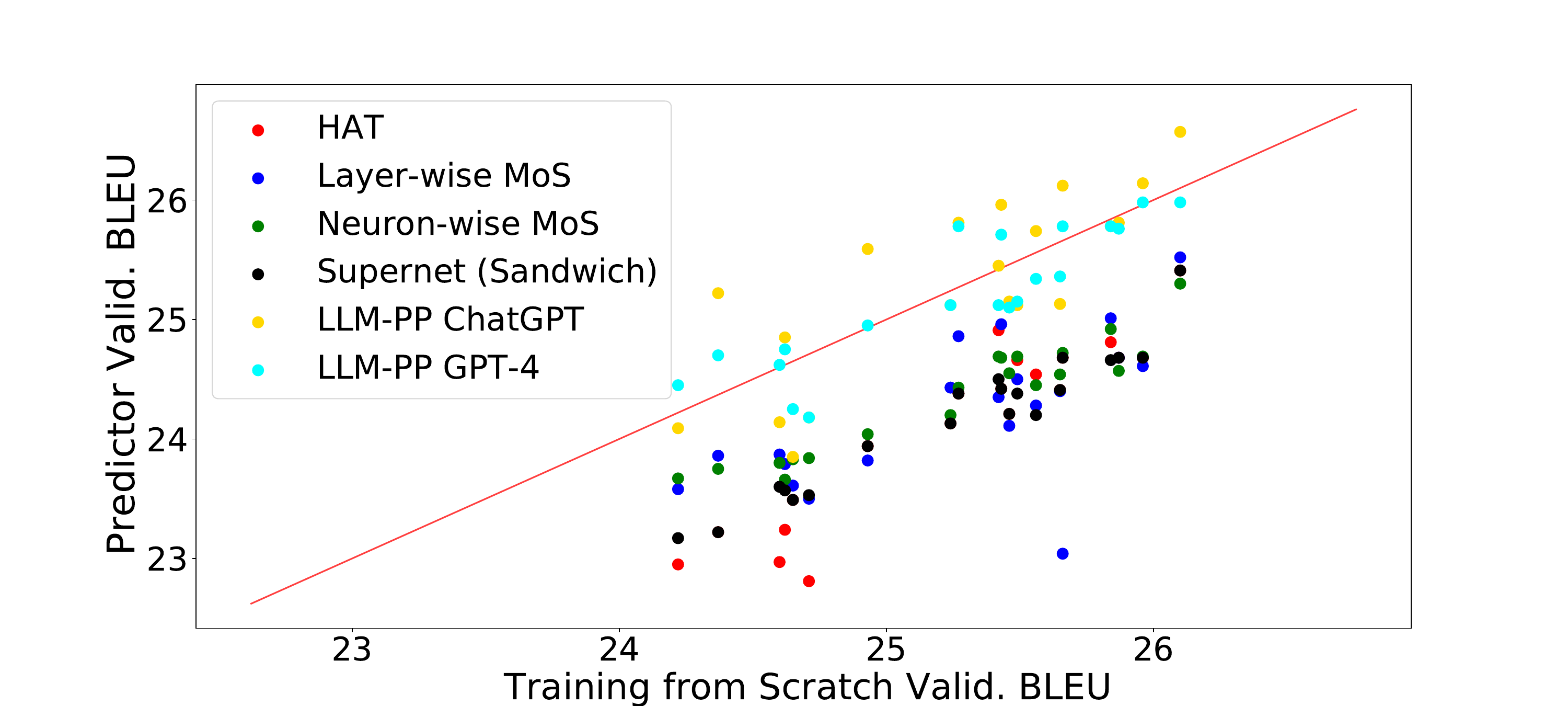}
        \caption{WMT'14 En-De}
    \end{subfigure}%
    ~ 
    \begin{subfigure}[t]{0.3\textwidth}
        \centering
        \includegraphics[height=1.3in, width=1.9in]{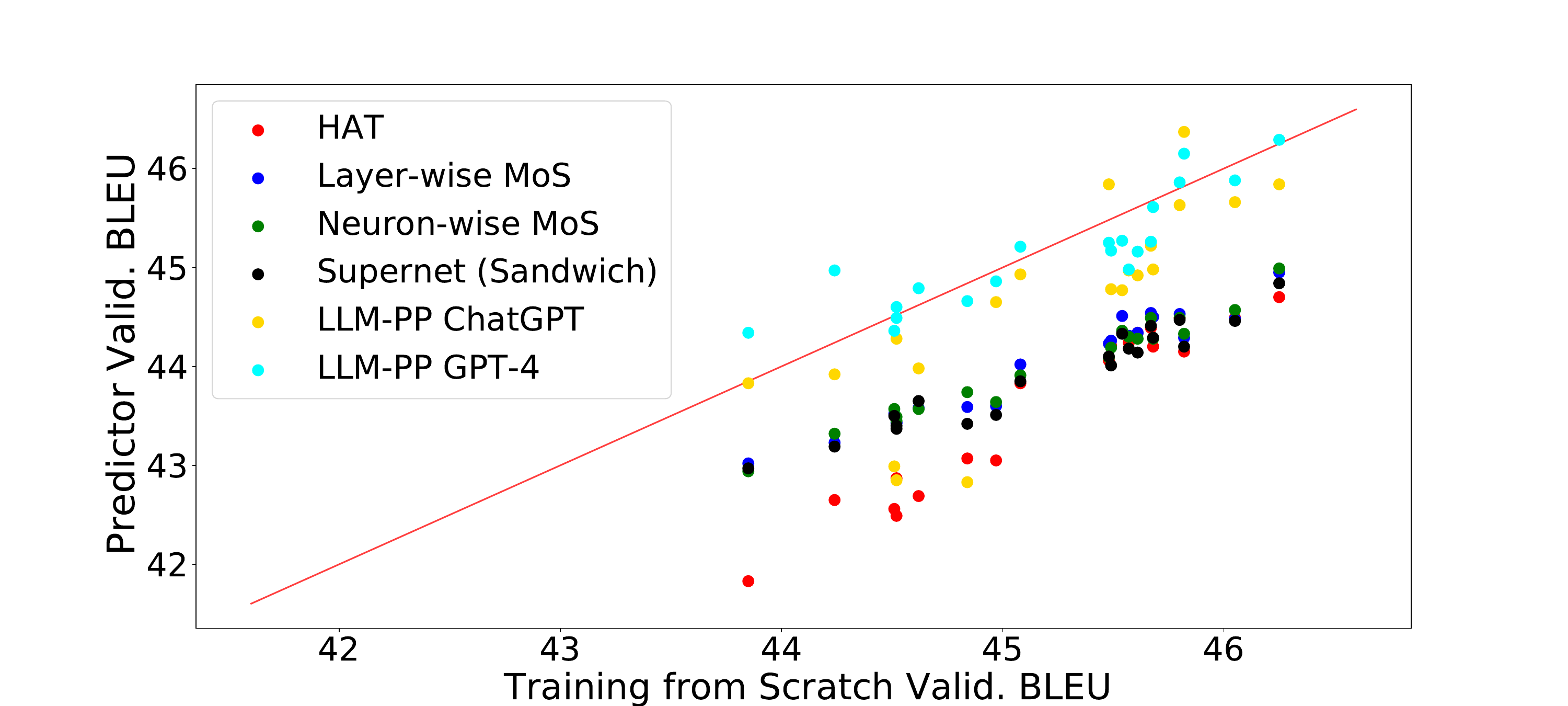}
        \caption{WMT'14 En-Fr}
    \end{subfigure}
    ~
    \begin{subfigure}[t]{0.3\textwidth}
        \centering
        \includegraphics[height=1.3in, width=1.9in]{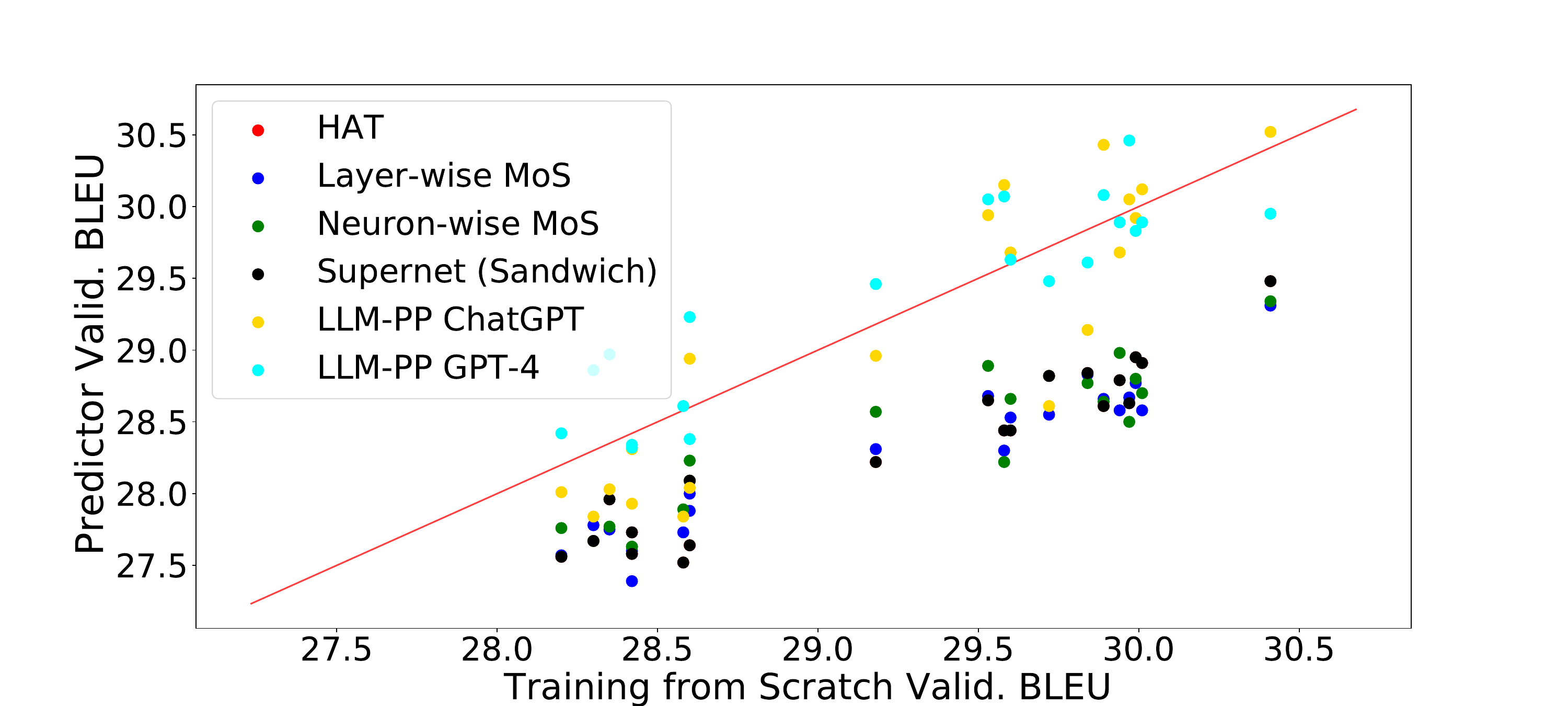}
        \caption{WMT'19 En-De}
    \end{subfigure}%
    \caption{Training from scratch validation BLEU vs. performance predictor validation BLEU for WMT benchmarks. Performance scores from the optimal predictor should lie on the diagonal (red line). LLM-PP predicted performance scores are largely closer to the diagonal than other predictors.}
\label{fig:tfs-scratch_figures}
\end{figure*}

\begin{table*}
\vspace{-0.1in}
\scriptsize
\begin{center}
\begin{tabular}
{lllllllll}
\toprule
\textbf{Dataset} & \multicolumn{2}{c}{\textbf{WMT'14 En-De}} & \multicolumn{2}{c}{\textbf{WMT'14 En-Fr}} & \multicolumn{2}{c}{\textbf{WMT'19 En-De}} & \multicolumn{2}{c}{\textbf{Average}} \\
\textbf{Performance Predictor} & \textbf{MAE } & \textbf{Kendall } & \textbf{MAE } & \textbf{Kendall } & \textbf{MAE } & \textbf{Kendall } & \textbf{MAE ($\downarrow$)} & \textbf{Kendall ($\uparrow$)} 
\\ \midrule
\multicolumn{9}{l}{\textbf{Baseline}}  \\
HAT  & 1.14 & 0.71   & 1.59   & 0.79 & 0.91  & 0.72  & 1.21  & 0.74     \\
Supernet (Sandwich) & 1.05   & \textbf{0.81}   & 1.27   & 0.78  & 0.91   & 0.72   & 1.08    & 0.77  \\
Layer-wise MoS & 0.97  & 0.56   & 1.16  & 0.79   & 0.96  & \textbf{0.74}  & 1.03    & 0.70   \\
Neuron-wise MoS & 0.87  & 0.79   & 1.18   & \textbf{0.87}  & 0.87   & 0.67   & 0.97  & \textbf{0.78}   \\ \midrule
\multicolumn{9}{l}{\textbf{LLM-PP}}  \\
\change{Mistral} & \change{0.73} & \change{0.22} & \change{0.60} & \change{0.34} & \change{0.92} & \change{0.18} & \change{0.75} & \change{0.25} \\
ChatGPT  & 0.42  & 0.52  & 0.82   & 0.61 & 0.72  & 0.56  & 0.65   & 0.56    \\
GPT-4 & 0.28  & 0.65  & \textbf{0.28}  & 0.75  & {0.32}  & 0.65  & \textbf{0.29}  & 0.68  \\ \midrule
\multicolumn{9}{l}{\textbf{LLM-PP GPT-4 Ablation}}  \\
Demonstraions only & 0.31  & 0.52  & 0.30 & 0.66  & 0.34   & 0.61   & 0.32   & 0.60  \\
+ Role + Hyp. & 0.27   & 0.53 & 0.32  & 0.71   & 0.32  & 0.67  & 0.30  & 0.64   \\
+ First instruction & 0.26  & 0.60   & 0.34  & 0.68  & 0.34  	& 0.58  & 0.31  & 0.62 \\
+ Second instruction & 0.27  & 0.60 	& 0.31  & 0.72   & 0.35   & 0.66 & 0.31   & 0.66 \\
+ Third instruction & 0.31   & 0.50 & 0.33   & 0.73   & \textbf{0.29}   & 0.67  & 0.31  & 0.63  \\
+ Fourth instruction & 0.25   & 0.63   & 0.32   & 0.65   & 0.33   & 0.71  & 0.30  & 0.66  \\
+ Fifth instruction & 0.28   & 0.65 & \textbf{0.28}   & 0.75   & {0.32} & 0.65   & \textbf{0.29}  & 0.68  \\ \midrule
\multicolumn{9}{l}{\textbf{LLM-Distill-PP}}  \\
ChatGPT  & 0.32 & 0.6  & 1.01   & 0.79   & 0.95  & 0.65   & 0.76   & 0.68    \\
GPT-4 & \textbf{0.22}  & 0.64  & 0.34 & 0.76   & 0.38  & 0.68  & 0.31  & 0.69  \\ 
\bottomrule
\end{tabular}
\caption{Average MAE and Kendall-Tau between the performance predictor performance and the TFS performance, across three different seeds. Based on the last two columns, the main takeaways are as follows: (1) LLM-PP achieves the state-of-the-art MAE score, with marginal degradation in Kendall-Tau over supernet based performance predictors. (2) For LLM-PP GPT-4, MAE is agnostic to the different instructions/components of the prompt template. (3) The second (downstream task  examples) and the fourth instruction (efficiency metric description) are crucial for achieving good Kendall-Tau. (4) LLM-PP predictions can be distilled to a MLP-based regressor (LLM-Distill-PP) with mostly similar or improved MAE and Kendall-Tau scores, thanks to regularization and  context specialization to performance prediction task.} 
\label{tab:llm_pe_mae_tau_comparison}
\end{center}
\vspace{-0.3in}
\end{table*}

\subsection{Results}
\label{sec:llm_pe_results}

\noindent\textbf{LLM-PP predictions closely align with TFS performance scores compared to the baselines.} Figure~\ref{fig:tfs-scratch_figures} illustrates the TFS versus performance predictor validation BLEU for different WMT benchmarks. The diagonal line (red line) represents the perfect predictor, where the predicted performance exactly matches the TFS score. The predictions from the supernet-based predictors (i.e., all non-LLM-based ones) are consistently underestimates of the TFS performance for all architectures across three benchmarks. In contrast, LLM-PP predictions are largely closer to the diagonal line, showcasing the high accuracy of LLM-PP.

\noindent\textbf{LLM-PP achieves SoTA MAE, slightly trailing baselines in Kendall-Tau.} Table~\ref{tab:llm_pe_mae_tau_comparison} displays the MAE and Kendall-Tau of baseline and LLM-PP predictors. Neuron-wise MoS stands out as the best baseline on average across datasets, boasting the lowest MAE and highest Kendall-Tau score. \change{LLM-PP Mistral outperforms supernet-based baselines in MAE for WMT'14 En-De and WMT'14 En-Fr tasks.} LLM-PP ChatGPT and LLM-PP GPT-4 surpass Neuron-wise MoS in MAE, with LLM-PP GPT-4 achieving the SoTA MAE score. However, LLM-PP slightly lags behind baselines in Kendall-Tau. In~\ref{sec:kendall_tau_finegrained_analysis}, we examine the histogram of distances between items in discordant pairs in the gold ranking for Neuron-wise MoS and LLM GPT-4. Discordant pairs of LLM-PP mostly cluster around the low gold ranking distances region, similar to Neuron-wise MoS, which shouldn't significantly impact PP use cases (as observed in Section~\ref{sec:search_results}). The resulting CDF of gold ranking distances for discordant pairs for LLM-PP GPT-4 and Neuron-wise MoS are very similar. These results indicate that PP prompts can effectively design accurate performance predictors. Within LLM-PP, GPT-4 outperforms ChatGPT on both metrics across datasets.

\noindent\textbf{LLM-PP benefits from all the components of PP prompts.} 
The last major row in Table~\ref{tab:llm_pe_mae_tau_comparison} displays the performance when ablating different components of PP prompts. LLM-PP's overall superior performance is attributed to having all PP prompt components together. Surprisingly, LLM-PP outperforms baselines in MAE even without any instructions (Demonstration only), showcasing the LLM's remarkable ability to grasp the performance prediction task based solely on demonstrations. While the MAE performance of different ablation variants is largely similar, there are differences in Kendall-Tau performance across variants. The second instruction (introducing downstream task-specific examples) and the fourth instruction (describing the efficiency metric) play crucial roles in achieving high Kendall-Tau for LLM-PP.

\noindent\textbf{LLM-PP requires at least 10 demonstrations to achieve SoTA MAE.} \finalchange{The performance predictor's quality notably improves with an increase in the number of demonstrations, as illustrated in Table~\ref{tab:impact_with_demos}. It becomes evident that LLM-PP requires at least 10 demonstrations to surpass the performance of SoTA supernet-based performance predictors in terms of MAE.}

\begin{table*}
\scriptsize
\begin{center}
\begin{tabular}
{ccccc}
\toprule
\textbf{Dataset} & \multicolumn{2}{c}{\textbf{WMT'14 En-Fr}} & \multicolumn{2}{c}{\textbf{WMT'19 En-De}} \\
\textbf{Supernet} & \textbf{MAE ($\downarrow$)} & \textbf{Kendall ($\uparrow$)} & \textbf{MAE ($\downarrow$)} & \textbf{Kendall ($\uparrow$)}
\\ \midrule
Neuron-wise MoS & 1.18 & \textbf{0.87} & 0.87 & \textbf{0.67} \\
LLM-PP ChatGPT (1 demonstration) & 2.81 & 0.55 & 1.72 & 0.56 \\
LLM-PP ChatGPT (3 demonstrations) & 2.28 & 0.67 & 1.12 & 0.57 \\
LLM-PP ChatGPT (10 demonstrations) & \textbf{0.82} & 0.61 & \textbf{0.72} & 0.56 \\
\bottomrule
\end{tabular}
\caption{\finalchange{Average MAE and Kendall-Tau between the performance predictor performance and the TFS performance, across three different seeds. LLM-PP requires at least 10 demonstrations to surpass the performance of state-of-the-art supernet-based performance predictors in MAE.}}
\label{tab:impact_with_demos}
\end{center}
\end{table*}

LLM-PP exceeds non-supernet baselines~\cite{{colin2022adeeperlook}}, with LLM-PP GPT-4 achieving a high Kendall Tau, as discussed in~\ref{sec:llmpp_vs_nonsupernet}. LLM-PP attains SoTA MAE and SoTA Kendall-Tau scores for low-resource/indigenous languages~\cite{ebrahimi-etal-2023-findings} (see~\ref{sec:llmpp_recent_datasets}) and uncommon evaluation metric (COMET~\cite{rei-etal-2022-comet}, see~\ref{sec:llmpp_comet_metric}). LLM-PP provides fairly robust performance predictions (see~\ref{sec:llmpp_robust_preds}).
\change{While LLM-PP excels in performance prediction quality, its cost scales linearly with the number of predictions. This cost can become prohibitive, especially for PP-heavy applications like NAS, where the number of predictions can reach several thousand.}

\section{Distillation of LLM-PP}
\label{sec:llm_distill_pp}

To illustrate the cost, let's consider the example of NAS run by HAT~\citep{hat} for a latency constraint on a given hardware, involving the evaluation of approximately 3,000 candidate architectures. As of August 2023, the pricing for GPT-4 is $0.03$\$ per 1K tokens. Assuming PP prompts consume about one-third of 1K tokens, the estimated cost per constraint on a given hardware would be around $30$\$ ($\frac{0.03  * 3000}{3}$). The total cost depends on the number of constraint types (e.g., latency, memory, FLOPs), values (e.g., 100ms, 200ms), and hardware options (e.g., Nvidia A100, Raspberry Pi). For instance, with three constraint types, five values for each constraint, and four target hardwares, the estimated cost could soar to approximately $1,800$\$ ($\frac{0.03  * 3000 * 3 * 5 * 4}{3}$) per downstream task.

To address this cost challenge, we propose LLM-Distill-PP, a cost-effective alternative trained on distilled outputs of LLM-PP. LLM-Distill-PP, a MLP based regressor, is trained using a distillation dataset for the PP task. This dataset is created by sampling architectures from the search space and recording the downstream task performance predicted by LLM-PP. 
LLM-Distill-PP is trained using architecture-specific hyperparameters as features and the distilled output as labels. Once trained, LLM-Distill-PP can predict the performance of unseen architectures for the given downstream task. If the number of distillation examples is small (e.g., $3,000$), the estimated cost to query LLM-PP will be approximately $30$\$ ($\frac{0.03  * 3000}{3}$). This one-time cost of LLM-Distill-PP is amortized across different constraint types, values, and hardwares (e.g., 60 search runs), leading to a substantial $98.3$\% reduction in cost (from $1,800$\$ to $30$\$). 
\change{LLM-Distill-PP achieves a superior efficiency-accuracy tradeoff, offering comparable accuracy to LLM-PP but with significantly faster prediction times (0.01s vs. 11.9s), as detailed in~\ref{sec:llmdistillpp_perf_time}.}

\noindent\textbf{Setup.} 
LLM-Distill-PP's architecture encoding details can be found in~\ref{sec:mt_arch_encoding}. The hyperparameters of its regression model, borrowed from HAT's latency predictor, include 3 hidden layers, 400 as the hidden dimension, 128 as the batch size, $\mbox{1e-5}$ as the learning rate, and $5000$ as the number of training steps. Distillation from LLM-PP uses only 3000 architecture examples for each downstream task.

\noindent\textbf{Results.} 
LLM-Distill-PP's results are summarized in the third major row of Table~\ref{tab:llm_pe_mae_tau_comparison}. Despite its simple model design, LLM-Distill-PP performs similarly or better than LLM-PP for both ChatGPT and GPT-4. In the case of ChatGPT, LLM-Distill-PP exhibits an average improvement of roughly 17\% in both MAE and Kendall-Tau over LLM-PP. For GPT-4, LLM-Distill-PP has a 7\% lower average MAE compared to LLM-PP while maintaining similar Kendall-Tau. Notably, LLM-Distill-PP achieves the SoTA MAE for the WMT'14 En-De task, outperforming LLM-PP by 20\%. \change{Two main factors contribute to this result. First, 
the smaller size of LLM-Distill-PP (a linear regression model with only 486K parameters) reduces the likelihood of overfitting compared to LLM-PP (an LLM with several billion parameters), resulting in better performance. Second, LLM-Distill-PP is a specialist model with parameters trained specifically for the performance prediction task using a few thousand examples. In contrast, LLM-PP is a generalist model that performs in-context learning with PP prompts and 10 demonstrations.}

%% file: tex/search_initializer.tex
\section{LLM-Distill-PP for Architecture Search}
\label{sec:llm_pe_arch_search}

\begin{table*}[t]
\scriptsize
\begin{center}
\begin{tabular}{lccccc}
\toprule
\textbf{Search Algorithm} & \textbf{BLEU ($\uparrow$)} & \textbf{Latency (ms) ($\downarrow$)} & \textbf{GFLOPs ($\downarrow$)} &  \textbf{Model Size (M) ($\downarrow$)} &  \textbf{Search Hours ($\downarrow$)} \\ \midrule
\textbf{WMT'14 En-De} \\
HAT & {27.9} & 102.0 & 3.0 & 64.4 & 1.09 \\
Layer-wise MoS & 27.8 & 100.4 & 3.08 & 64.4 & 1.45 \\
Neuron-wise MoS & \textbf{28.0} & \textbf{99.0} & 3.26 & 72.2 & 1.39 \\
HS-NAS (GPT-4, HAT, 1, 15) & {27.9} & {99.7} & \textbf{2.96} & \textbf{63.1} & \textbf{0.56} \\ \midrule
\textbf{WMT'14 En-Fr} \\
HAT & 40.8 & \textbf{96.4} & 2.61 & \textbf{63.8} & 6.33 \\
Layer-wise MoS & 40.5 & 99.4 & 2.96 & 70.5 & 6.81 \\
Neuron-wise MoS & \textbf{40.9} & 97.6 & 3.13 & 70.5 & 7.03 \\
HS-NAS (GPT-4, HAT, 1, 15) & {40.7} & {98.2} & \textbf{2.54} & \textbf{63.8} & \textbf{3.15} \\ \midrule
\textbf{WMT'19 En-De} \\
HAT & 44.7 & 100.8 & 3 & 73.06 & 1.11 \\
Layer-wise MoS & \textbf{44.9} & 96.8 & 3.26 & 82.95 & 1.13  \\
Neuron-wise MoS & \textbf{44.9} & 122.4 & 3.34 & 82.95 & 1.21 \\
HS-NAS (GPT-4, HAT, 1, 15) & 44.4 & \textbf{70.0} & \textbf{2.51} & \textbf{66.36} & \textbf{0.46} \\
\bottomrule
\end{tabular}
\caption{HS-NAS versus SoTA NAS on three MT benchmarks for latency constraint of $100ms$ - Test BLEU, latency in milliseconds, GFLOPs, model size in millions, and search hours.}
\label{tab:nas_main_results_across_datasets}
\end{center}
\end{table*}

Given LLM-Distill-PP's ability to achieve high-performance prediction quality in a cost-effective manner, we explore its application in a real-world task: NAS. In NAS, performance predictors typically rank candidate architectures to identify high-performing ones. As discussed in Section~\ref{sec:related_work}, existing NAS research in NLP primarily uses weight-sharing supernets as performance predictors. Therefore, we address the research question: \textit{Can LLM-Distill-PP accelerate architecture search while maintaining the efficiency and quality of SoTA NAS?} To answer this question, we introduce the Hybrid-Search algorithm for NAS (HS-NAS).
The core idea of HS-NAS is to employ LLM-Distill-PP for a subset of search iterations, utilizing the supernet for the remaining iterations. This approach is applied to the evolutionary search algorithm proposed in HAT.
\begin{algorithm}[h!]
\textbf{Input:} LLM-Distill-PP model: $\texttt{\mbox{llm-distill-pp}}$, Weight-sharing supernet: $\texttt{\mbox{supernet}}$, Latency predictor: $\texttt{\mbox{latency-predictor}}$, \#Search iterations: $\texttt{\mbox{num-iterations}}$,  Population size: $\texttt{\mbox{population-size}}$, 
Latency constraint: $\texttt{\mbox{latency-constraint}}$, LLM-Distill-PP Start Iteration: $\texttt{\mbox{llm-start-iteration}}$,  LLM-Distill-PP End Iteration: $\texttt{\mbox{llm-end-iteration}}$, ...  \\
\textbf{Output:} $\texttt{\mbox{best-architecture}}$ 
     \begin{algorithmic}[1]
     \State $popu \gets$ $\texttt{\mbox{population-size}}$ rand. samples from  search space {\textcolor{blue}{// create init. population}}
     \For{$iter \gets 1 \: \texttt{\mbox{to}} \: \texttt{\mbox{num-iterations}}$}
        \State {\textcolor{blue}{// gen. parents by picking top cand. arch.}}
        \If{{\textcolor{red}{ $\texttt{\mbox{llm-start-iteration}} < iter < \texttt{\mbox{llm-end-iteration}} $}} }
          \State {\textcolor{red}{$\texttt{\mbox{parents}} \gets$ top `$\texttt{\mbox{num-parents}}$' arch. from $popu$ by $\texttt{\mbox{llm-distill-pp}}$ }}
        \Else
          \State {\textcolor{red}{$\texttt{\mbox{parents}} \gets$ top `$\texttt{\mbox{num-parents}}$' arch. from $popu$ by $\texttt{\mbox{supernet}}$ }}
        \EndIf
        \State $\texttt{\mbox{mut-popu}}$ = HAT's mutation logic
        \State $\texttt{\mbox{cross-popu}}$ = HAT's crossover logic
        \State $popu = \texttt{\mbox{parents}} \cup \texttt{\mbox{mut-pop}} \cup \texttt{\mbox{cross-pop}}$ 
     \EndFor
     \State {{return top arch. from $popu$}}
\end{algorithmic}
\caption{Hybrid-Search algorithm for Neural Architecture Search (HS-NAS). Changes to HAT's search algorithm are in {\textcolor{red}{red}} color. The expanded algorithm can be found in~\ref{sec:hsnas_algo}.}
\label{algo:hsnasalgo}
\end{algorithm}

\begin{table*}
\scriptsize
\begin{center}
\begin{tabular}{lccccc}
\toprule
\textbf{Search  Algorithm} & \textbf{BLEU ($\uparrow$)} & \textbf{Latency (ms) ($\downarrow$)} & \textbf{GFLOPs ($\downarrow$)} &  \textbf{Model Size (M) ($\downarrow$)} &  \textbf{Search Hours ($\downarrow$)} \\ \midrule
$\mathbf{100ms}$ \\
HAT & 40.8 & \textbf{96.4} & 2.61 & \textbf{63.8} & 6.33 \\
Layer-wise MoS & 40.5 & 99.4 & 2.96 & 70.5 & 6.81 \\
Neuron-wise MoS & \textbf{40.9} & 97.6 & 3.13 & 70.5 & 7.03 \\
HS-NAS (GPT-4, HAT, 1, 15) & {40.7} & {98.2} & \textbf{2.54} & \textbf{63.8} & \textbf{3.15} \\ \midrule
$\mathbf{150ms}$  \\
HAT & 41.3 & 176.4 & \textbf{3.31} & \textbf{74.3} & 7.33 \\
Layer-wise MoS & \textbf{41.4} & \textbf{158.7} & 4.3 & 92.8 & 8.39 \\
Neuron-wise MoS & \textbf{41.4} & 200.2 & 4.26 & 92.8 & 8.35 \\
HS-NAS (GPT-4, HAT, 1, 15) & \textbf{41.4} & {172.6} & \textbf{3.31} & \textbf{74.3} & \textbf{3.69}  \\ \midrule
$\mathbf{200ms}$  \\
HAT & 41.5 & 187.5 & \textbf{3.7} & \textbf{79.5} & 7.8 \\
Layer-wise MoS & 41.4 & 205.6 & 4.49 & 99.4 & 8.63 \\
Neuron-wise MoS & 41.6 & \textbf{184.1} & 4.53 & 99.4 & 8.77 \\
HS-NAS (GPT-4, HAT, 1, 15) & \textbf{42.0} & 187.8 & \textbf{3.7} & \textbf{79.5} & \textbf{3.88} \\
\bottomrule
\end{tabular}
\caption{HS-NAS versus SoTA NAS on WMT'14 En-Fr for different latency constraints - Test BLEU, latency in milliseconds, GFLOPs, model size in millions, and search hours. }
\label{tab:nas_main_results_across_constraints}
\end{center}
\end{table*}

\begin{table*}
\scriptsize
\begin{center}
\begin{tabular}{lccccc}
\toprule
\textbf{Search  Algorithm} & \textbf{BLEU ($\uparrow$)} & \textbf{Latency (ms) ($\downarrow$)} & \textbf{GFLOPs ($\downarrow$)} &  \textbf{Model Size (M) ($\downarrow$)} &  \textbf{Search Hours ($\downarrow$)} \\ \midrule
HAT & {27.9} & 102.0 & 3.0 & 64.4 & 1.09 \\
HS-NAS (GPT-4, HAT, 1, 30) & 27.5 & 99.3 & 3.34 & 72.2 & 0.04 \\
HS-NAS (GPT-4, HAT, 1, 5) & 27.4 & 100.4 & 2.96 & 63.1 & 0.97 \\
HS-NAS (GPT-4, HAT, 25, 30) & 28.0 & 119.1 & 3.18 & 70.9 & 0.95 \\
HS-NAS (GPT-4, HAT, 1, 15) & \textbf{27.9} & \textbf{99.7} & \textbf{2.96} & \textbf{63.1} & \textbf{0.56} \\
HS-NAS (GPT-4, HAT, 16, 30) & 27.6 & 101.7 & 3.34 & 72.2 & 0.75 \\
HS-NAS (GPT-4, HAT, 1, 25) & 27.7 & 98.9 & 3.01 & 63.1 & 0.23 \\
\bottomrule
\end{tabular}
\caption{HS-NAS versus HAT on WMT'14 En-De for latency constraint: $100ms$ - Test BLEU, latency in milliseconds, GFLOPs, model size in millions, and search hours.}
\label{tab:nas_main_results_across_schedules}
\end{center}
\end{table*}

LLM-Distill-PP will be used as performance predictor for all the search iterations in between $\texttt{\mbox{llm-start-iteration}}$ and $\texttt{\mbox{llm-end-iteration}}$. In rest of the iterations, supernet will be used as performance predictor. When $\texttt{\mbox{llm-start-iteration=1}}$ and $\texttt{\mbox{llm-end-iteration=num-iterations}}$, HS-NAS uses LLM-Distill-PP as performance predictor for all the search iterations. 
HS-NAS comes with four arguments: ($\texttt{\mbox{llm-distill-pp}}$, $\texttt{\mbox{supernet}}$, $\texttt{\mbox{llm-start-iteration}}$, $\texttt{\mbox{llm-end-iteration}}$). 
For all our search experiments, we use LLM-Distill-PP GPT-4 as $\texttt{\mbox{llm-distill-pp}}$ due to its superior performance over the ChatGPT counterpart (see the third major row in Table~\ref{tab:llm_pe_mae_tau_comparison}). We use the $\texttt{\mbox{latency-predictor}}$ and $\texttt{\mbox{supernet}}$ from HAT. 
Other details of the setup (e.g., efficiency metric for search (search hours), and architecture (latency, GFLOPs, model size)) can be seen in~\ref{sec:search_and_eval_details}.


\subsection{Results}
\label{sec:search_results}

\noindent\textbf{Varying benchmarks.} 
HS-NAS shows comparable performance to the SoTA across benchmarks, achieving approximately a 50\% reduction in search hours. In some cases, it even enhances latency, GFLOPs, and model size, as illustrated in Table~\ref{tab:nas_main_results_across_datasets}. This pattern highlights the effectiveness of using LLMs as good initializers for architecture search.

\noindent\textbf{Varying latency constraints.} 
The trend observed in HS-NAS remains consistent across different latency constraints. Table~\ref{tab:nas_main_results_across_constraints} presents a comparison of the HS-NAS configuration (GPT-4, HAT, 1, 15) against the SoTA NAS for different latency constraints: 100ms, 150ms, and 200ms. Alongside a 50\% reduction in search hours, HS-NAS attains comparable or improved GFLOPs and maintains the same model size compared to SoTA NAS.

\noindent\textbf{Varying start and end iteration pairs.} 
Among different start and end iteration pairs, HS-NAS utilizing LLM-Distill-PP (GPT-4) for the initial 50\% of iterations and HAT supernet for the remainder performs comparably or outperforms HAT across all metrics. Table~\ref{tab:nas_main_results_across_schedules} presents the results of HS-NAS for various start and end iteration pairs. Utilizing LLM-Distill-PP for the entire search yields lower performance, indicating that a marginal degradation in Kendall-Tau hinders LLM-Distill-PP's effectiveness in handling the complete search. These trends underscore the utility of a predictor with SoTA MAE scores for the initial search, while a predictor with SoTA Kendall-Tau is valuable for the later stages of the search.

\noindent\textbf{Varying initialization seeds, FLOPs constraints, underlying supernet.}
HS-NAS exhibits resilience to initialization effects stemming from different seeds, yielding largely consistent results across metrics. Further details are provided in~\ref{sec:var_init_seeds}. HS-NAS performs comparably to HAT under varying FLOPs constraints, showcasing a minimum 16\% reduction in search hours, a 1.2\% improvement in latency, consistent GFLOPs, and identical model sizes. These trends persist consistently across benchmarks, as outlined in~\ref{sec:var_flop_constraints}. The superiority of HS-NAS remains robust across different underlying supernets, as elucidated in~\ref{sec:var_underlying_supernet}.

\noindent\textbf{Trivially constructed efficient adaptations of SoTA.} Search hours can be trivially reduced in several ways: halving the total number of search iterations and/or using distilled SoTA predictor instead of using supernet predictor directly. 
While these adaptations lead to a big drop in BLEU performance (1.8\% for HAT ($\texttt{\mbox{num-iter.}}$=15)) or a big increase in latency and GFLOPs (9.7\% and 32\% respectively for Distilled HAT ($\texttt{\mbox{num-iter.}}$=15)), HS-NAS dominates these adaptions in search hour reductions, while maintaining SoTA performance and not degrading on any footprint metric, as detailed in~\ref{sec:efficient_adaptations}.

Putting all the observed trends of HS-NAS together, we find that the generality of HS-NAS extends to constraint types (latency, FLOPs), constraint values (different latencies, different FLOPs),  different tasks (MT benchmarks), and underlying supernet (HAT, Neuron-wise MoS), while being robust to initialization effects.

%% file: tex/conclusion.tex
\section{Conclusion}
\label{conclusion}

This work shows that LLMs can be employed to create accurate and cost-effective performance predictors, providing insights into enhancing NAS. This contribution adds to the expanding field of LLMs in NAS, suggesting future research directions in adapting LLMs for both candidate architecture generation and joint performance prediction.


%% file: tex/limitations.tex
\section{Limitations}
\label{sec:c5_limitations}
\begin{itemize}
\item \change{\textbf{Expanding task domains.} Our evaluation setup, centered on machine translation benchmarks, aligns with existing NAS for NLP literature~\cite{hat,jawahar-etal-2023-automoe,jawahar2023mixtureofsupernets}, primarily focusing on machine translation tasks. Investigating the applicability of the LLM-PP framework to diverse NLP tasks (e.g., summarization, language modeling) and non-NLP domains (e.g., speech recognition, computer vision) stands as a crucial avenue for future exploration.}
\item \change{\textbf{Exploring diverse architectures.} This work 
focused on classic Transformer architectures as outlined by \citeauthor{transformer}, aligning with NAS for NLP literature. While our primary investigation remained focused on these architectures, examining other architecture types (e.g., convolution embedding based~\cite{salesky-etal-2023-multilingual}) stands as a pertinent future direction.}
\end{itemize}

\section*{Acknowledgments}
\label{sec:ack}
\finalchange{MAM acknowledges support from Canada Research Chairs (CRC), the Natural Sciences and Engineering Research Council of Canada (NSERC; RGPIN-2018-04267), Canadian Foundation for Innovation (CFI; 37771), and Digital Research Alliance of Canada.\footnote{\href{https://alliancecan.ca}{https://alliancecan.ca}} Lakshmanan's research was supported in part by a grant from NSERC (Canada).} 
We used ChatGPT for rephrasing and grammar checking of the paper.

%% file: tex/appendix.tex
\section{Appendix}
\label{sec:appendix}


\subsection{Related Work - Extended}
\label{sec:related_work_extended}

\noindent\textbf{LLMs.} LLMs can be classified into two categories based on their training methods: foundation and instruction-tuned LLMs. Foundation LLMs, which includes GPT-3~\citep{gpt3}, GLaM~\citep{glam}, LLaMA-1~\citep{llama1}, undergo language model training on unannotated corpus from the web. %
These LLMs typically encode a lot of useful knowledge in their parameters 
and can be used for a downstream task by either fine-tuning or zero/few-shot prompting. Instruction-tuned LLMs are usually foundation LLMs that undergo instruction-tuning, where LLMs are explicitly fine-tuned to follow user defined instructions well. Such LLMs include  InstructGPT~\citep{ouyang2022training}, ChatGPT~\citep{chatgpt}, 
GPT-4~\citep{gpt4}, LLaMA-2~\citep{llama2}, and PaLM-2~\citep{anil2023palm}. In practice, instruction-tuned LLMs can follow a wide range of user's instructions, even those that are outside the instruction tuning data distribution~\citep{ouyang2022training}. However, depending on the task, instruction-tuned LLMs are prone to generating content that are factually incorrect, hallucinated, ignores user's instruction, toxic, and so on~\citep{ouyang2022training}. These challenges make the current SoTA LLMs unreliable for critical applications such as medical diagnosis~\citep{singhal2022large}.

\noindent\textbf{Distilling LLMs.} Distilling the generations from LLMs to smaller student models has become commonplace in NLP these days~\citep{alpaca,vicuna2023,lamini-lm,mukherjee2023orca}. 
The key motivations for such efforts include: (i) \textit{cost reduction}: most LLMs are either behind a paywall or require high-end GPUs (e.g., NVIDIA A100) with high GPU memory (e.g., 80GB) to use, (ii) \textit{latency reduction}: most LLMs are too slow even on high-end  hardware (e.g., OPT-175B takes 4s for decoding 16 sequences of length 1024 on 8 NVIDIA A100 80GB GPUs~\citep{xiao2022smoothquant}), and (iii) \textit{customization}: most LLMs are general purpose and are difficult to finetune. The commonly used distillation technique is sequence level knowledge distillation~\citep{kim-rush-2016-sequence}, where the student models are finetuned on responses from teacher LLMs via a standard language modeling objective.

\subsection{Examples for Metrics}
\label{sec:metrics_examples}

\subsubsection{Mean Absolute Error}
\change{If predictions and TFS performances match perfectly, MAE will be zero, e.g., predictions are [23.4, 25.9, 28.1] and TFS performances are [23.4, 25.9, 28.1]. If predictions and TFS performances are mostly similar, MAE will be  low, e.g., predictions are [23.4, 25.9, 28.1] and TFS performances are [23.3, 25.8, 28.2], MAE is 0.1. If predictions and TFS performances are extremely different, MAE will be high, e.g., predictions are [21.2, 24.0, 22.1] and TFS performances are [23.3, 25.8, 28.2], MAE is 3.33.}

\subsubsection{Kendall-Tau}
\change{If predictions and TFS performances match perfectly, Kendall-Tau will be 100, e.g., predictions are [23.4, 25.9, 28.1] and TFS performances are [23.4, 25.9, 28.1]. If predictions and TFS performances are different but their architecture rankings are similar, Kendall-Tau will be 100, e.g., predictions are [23.4, 25.9, 28.1] and TFS performances are [22.2, 23.4, 25.1]. If predictions and TFS performances are different and their architecture rankings are dissimilar, Kendall-Tau will be negative, e.g., predictions are [23.4, 25.9, 28.1] and TFS performances are [23.4, 25.1, 22.2], Kendall-Tau is -0.33.}

\subsection{Prompt Template - Expanded version}
\label{sec:prompt_template_expanded}
\begin{figure*}
\centering
\includegraphics[width=4.5in,height=8.5in]{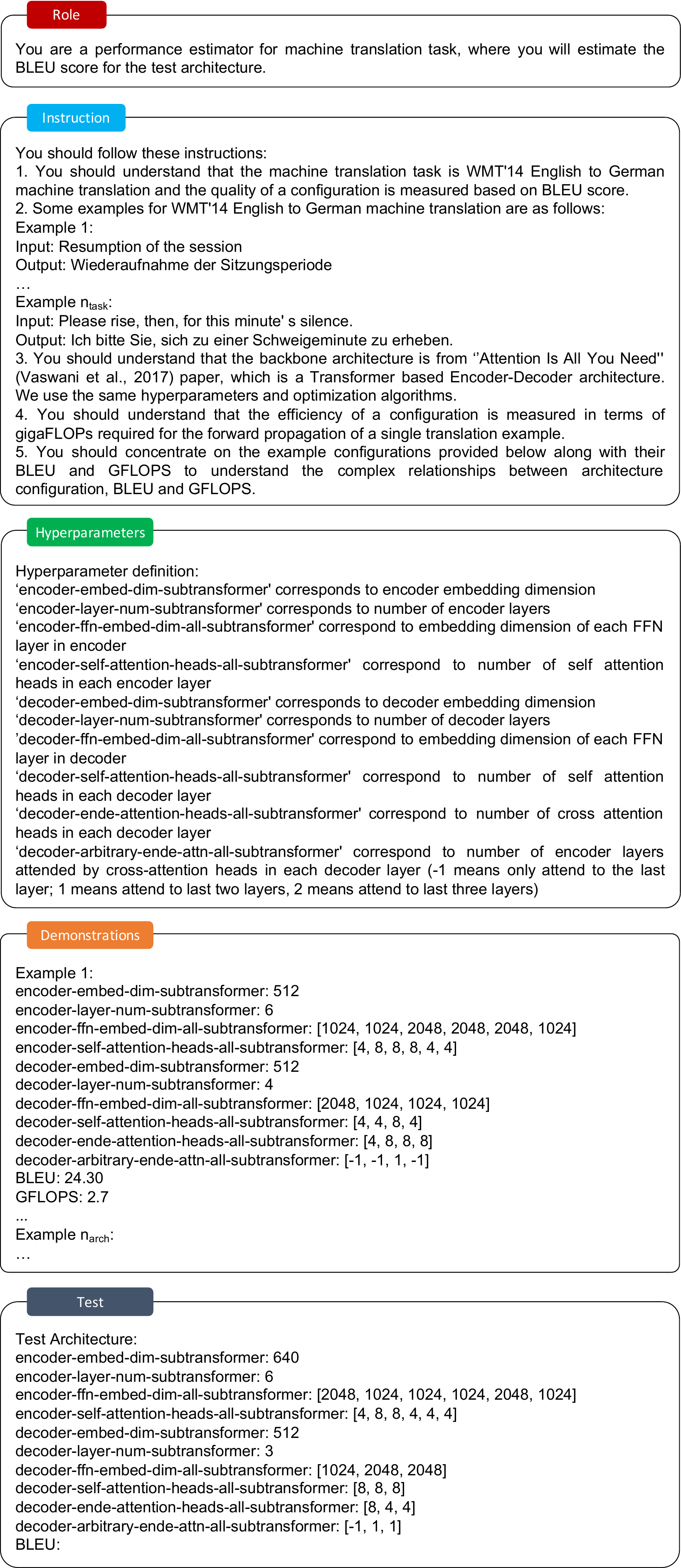}
\caption{Prompt template to prompt LLM to generate performance predictions for WMT'14 EN-DE task.}
\label{fig:framework_long}
\end{figure*}

The expanded version of the prompt template can be seen in  Figure~\ref{fig:framework_long}.

\subsection{Prompt Template - Design Process}
\label{sec:prompt_template_design_process}
\change{The design process began by examining crucial elements of the machine translation task, commonly used model architectures, and relevant efficiency metrics. Initially, we presented only {\textcolor{orange}{\textit{demonstrations}}}, borrowing hyperparameter wording from HAT’s configuration file. Subsequently, we added the {\textcolor{red}{\textit{role}}} and definition of each {\textcolor{green}{\textit{hyperparameters}}}, using wording from HAT’s helper description. Moving forward, our aim was to craft instructions enabling the LLM to grasp essential tasks, architecture, and metric details. Most instructions are prefixed with `You should' to encourage strict adherence. Five instructions were incorporated. The first specifies the dataset, translation direction, and quality metric. The second provides examples randomly sampled from the training set, presented with generic prefixes (`Input:' for source sentence, `Output:' for target sentence). The third outlines the architecture, citing the `Attention Is All You Need'~\cite{transformer} paper, assuming the LLM is familiar with this popular work. Standard settings and optimization algorithms are noted for training the architectures. The fourth identifies the efficiency metric in the demonstrations. The final instruction aims to summarize the relationships the LLM should learn to solve the task effectively.}

\subsection{Kendall-Tau - Fine-grained analysis}
\label{sec:kendall_tau_finegrained_analysis}
\begin{figure*}[t!]
    \centering
    \begin{subfigure}[t]{0.32\textwidth}
        \centering
        \includegraphics[height=1.6in, width=2.0in]{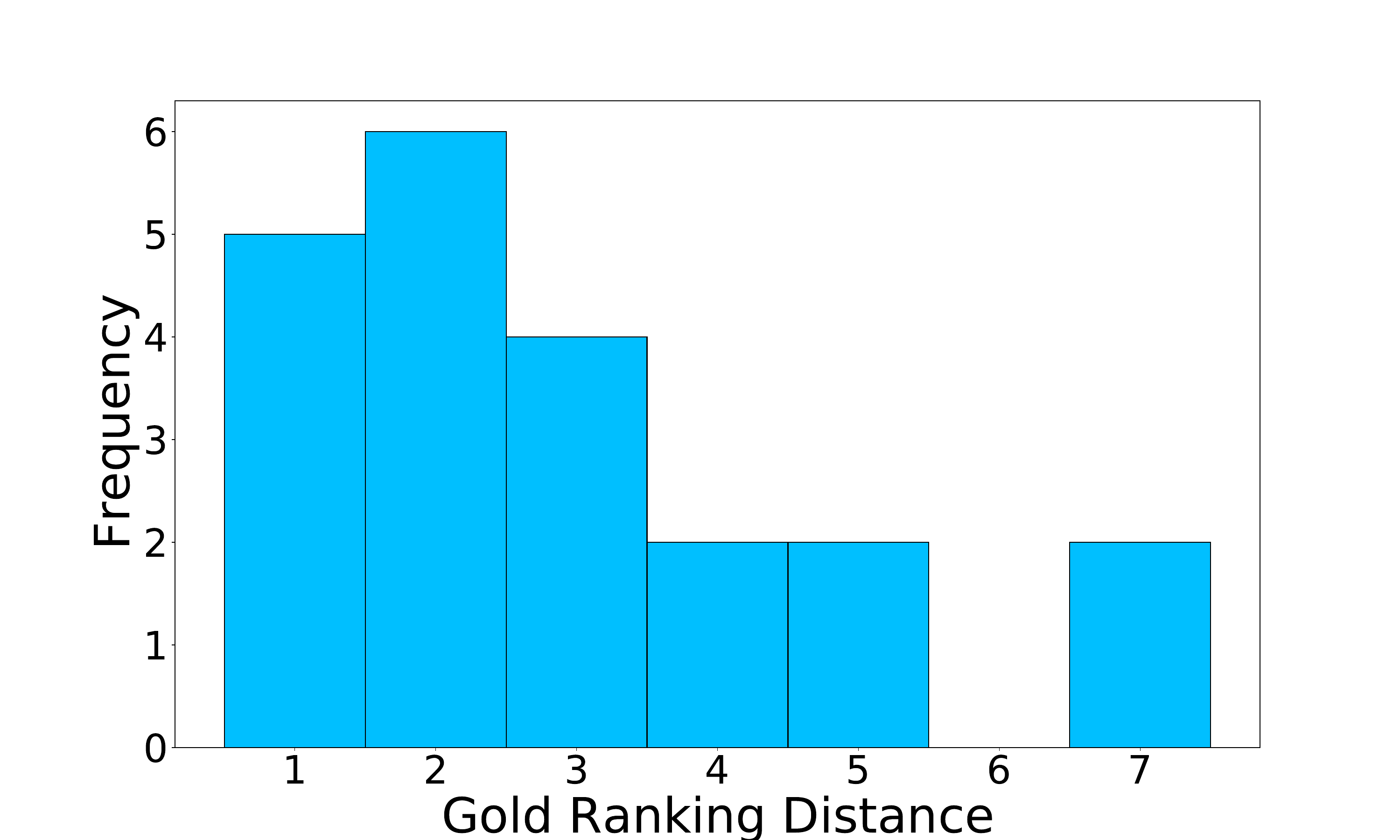}
        \caption{Neur. MoS - WMT'14 En-De}
    \end{subfigure}%
    ~ 
    \begin{subfigure}[t]{0.32\textwidth}
        \centering
        \includegraphics[height=1.6in, width=2.0in]{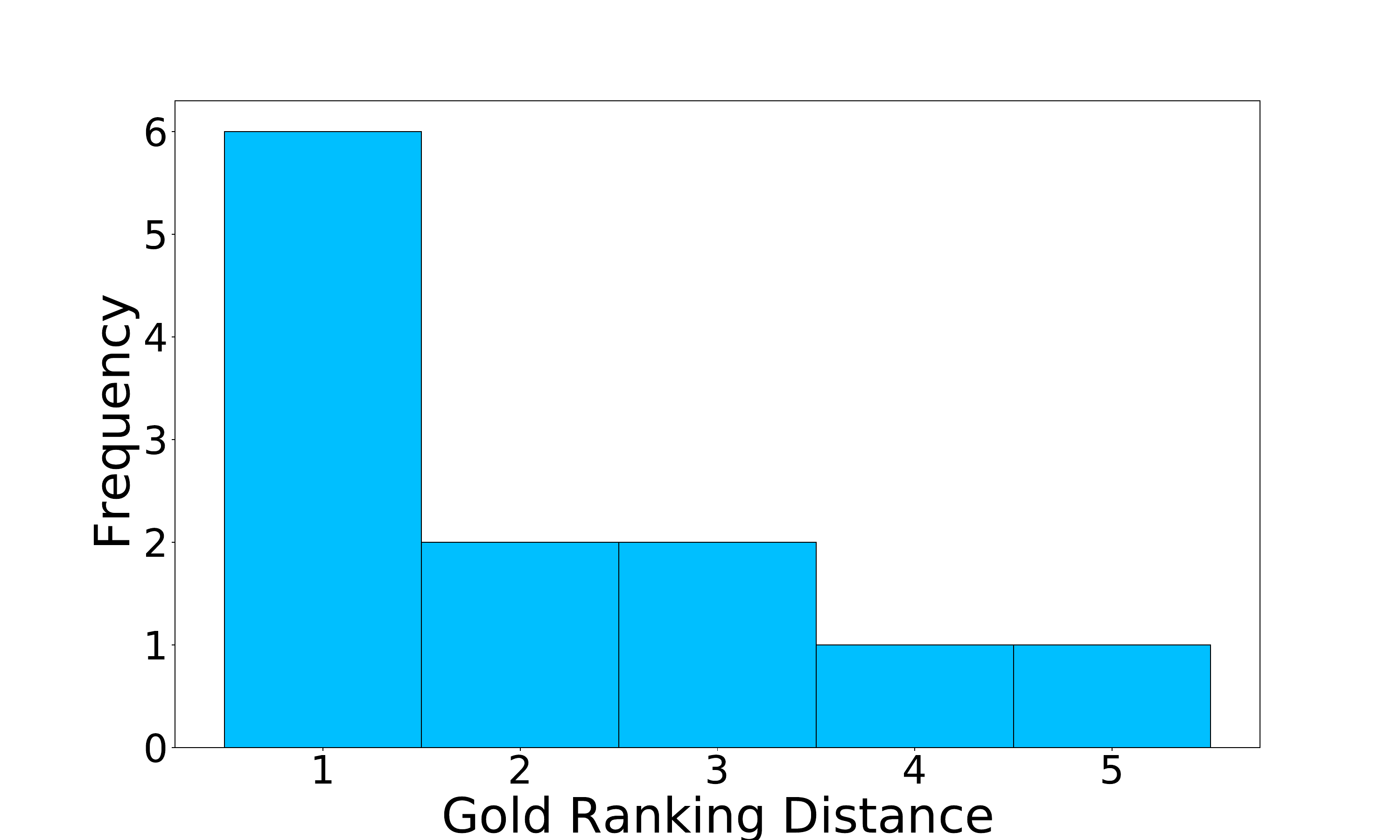}
        \caption{Neur. MoS - WMT'14 En-Fr}
    \end{subfigure}
    ~
    \begin{subfigure}[t]{0.32\textwidth}
        \centering
        \includegraphics[height=1.6in, width=2.0in]{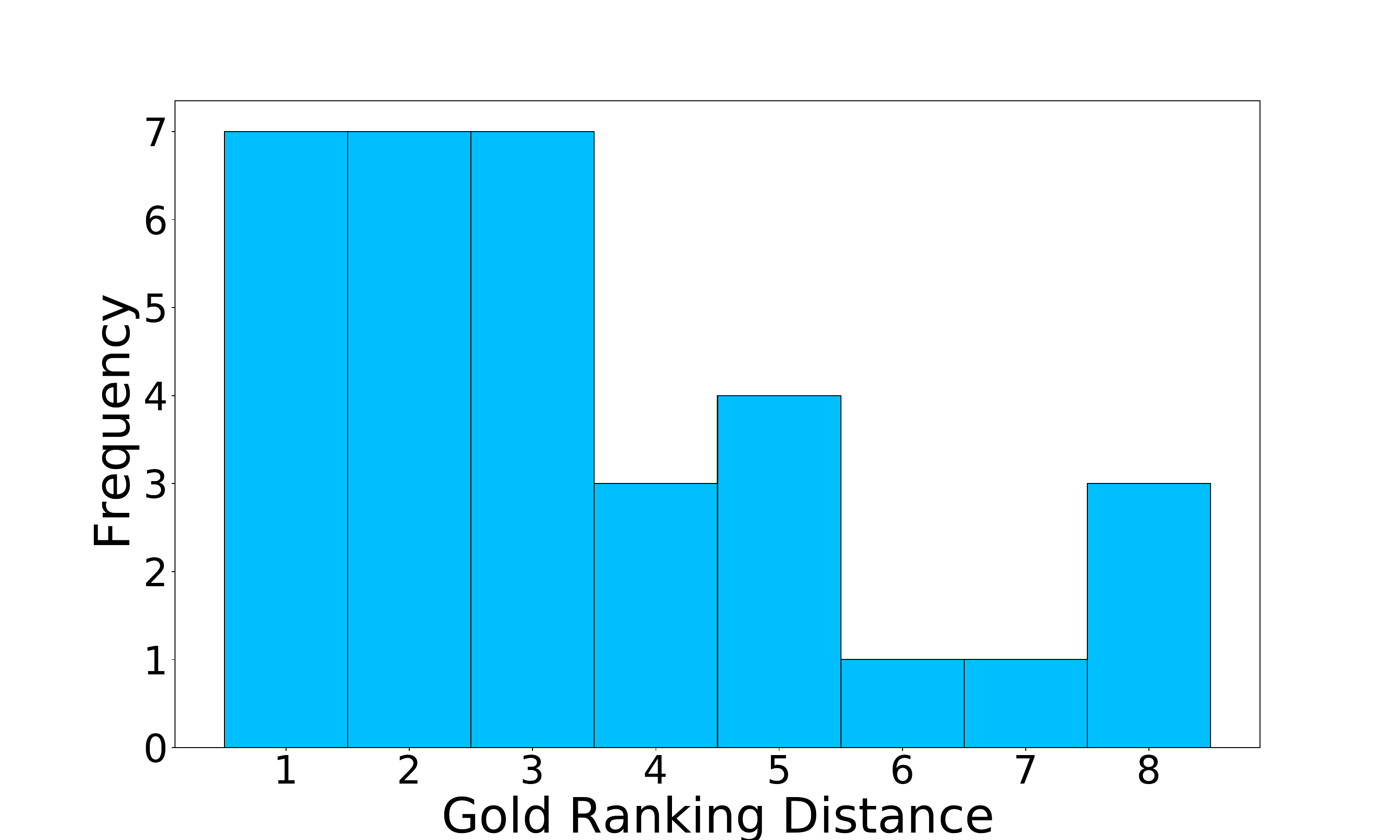}
        \caption{Neur. MoS - WMT'19 En-De}
    \end{subfigure}
    \begin{subfigure}[t]{0.32\textwidth}
        \centering
        \includegraphics[height=1.6in, width=2.0in]{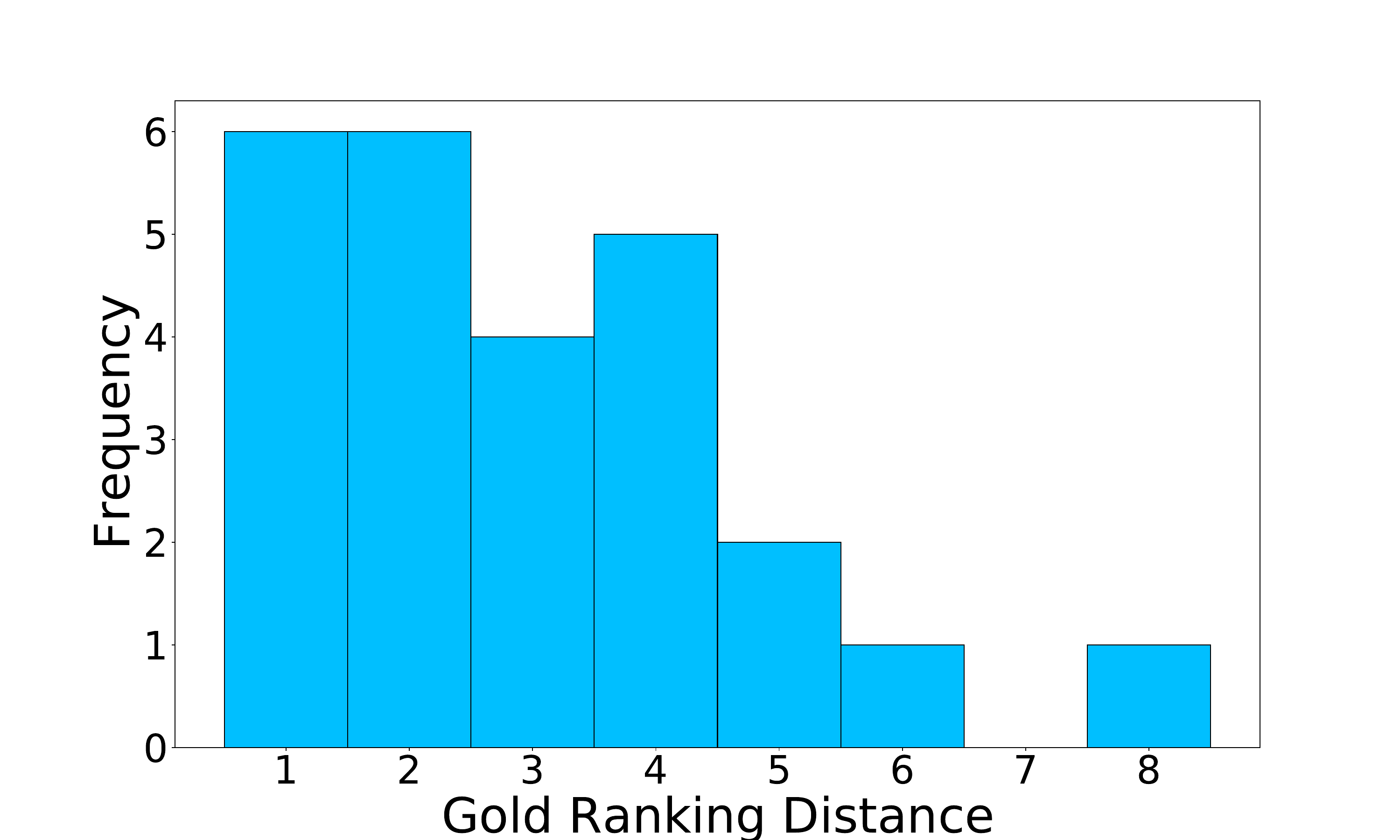}
        \caption{LLM GPT-4 - WMT'14 En-De}
    \end{subfigure}%
    ~ 
    \begin{subfigure}[t]{0.32\textwidth}
        \centering
        \includegraphics[height=1.6in, width=2.0in]{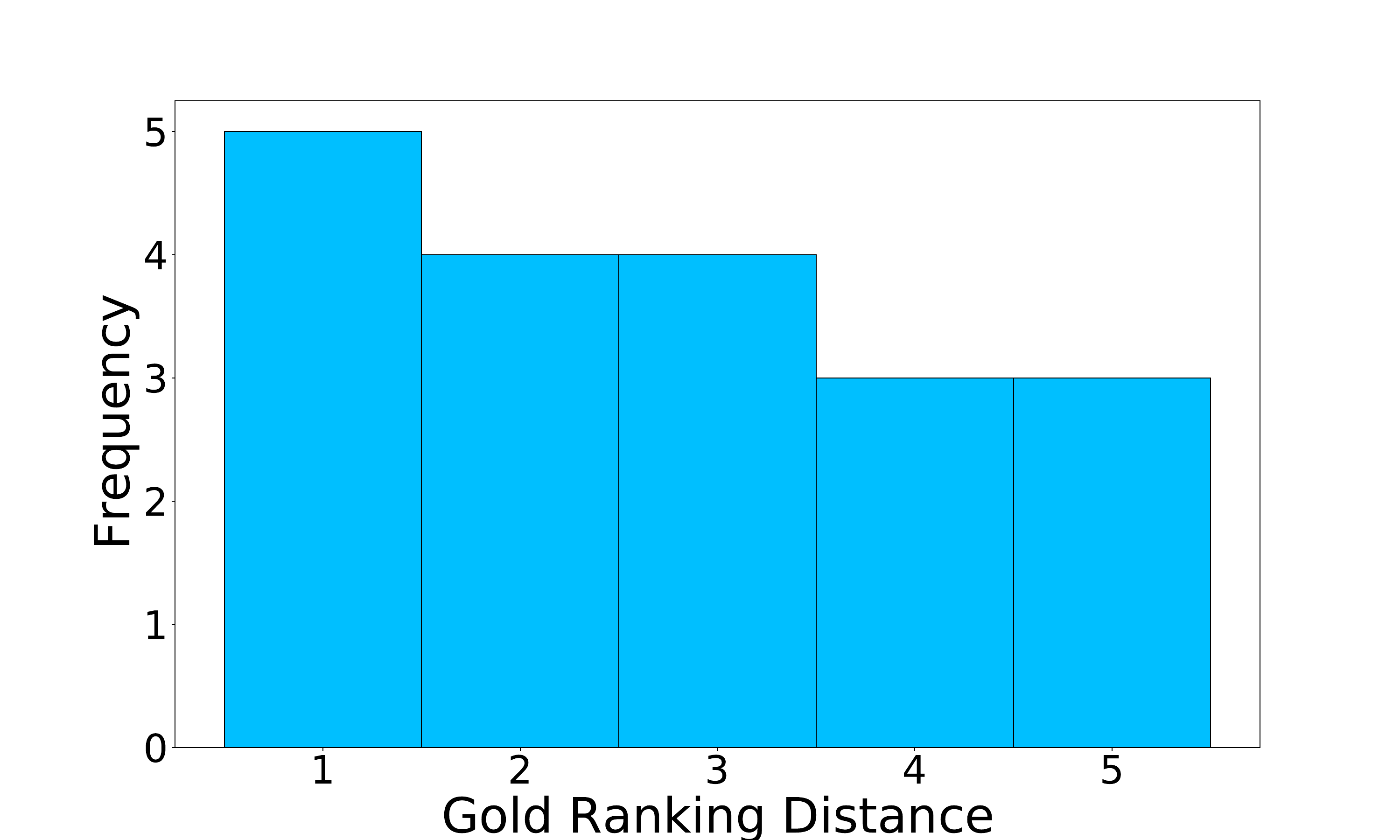}
        \caption{LLM GPT-4 - WMT'14 En-Fr}
    \end{subfigure}
    ~
    \begin{subfigure}[t]{0.32\textwidth}
        \centering
        \includegraphics[height=1.6in, width=2.0in]{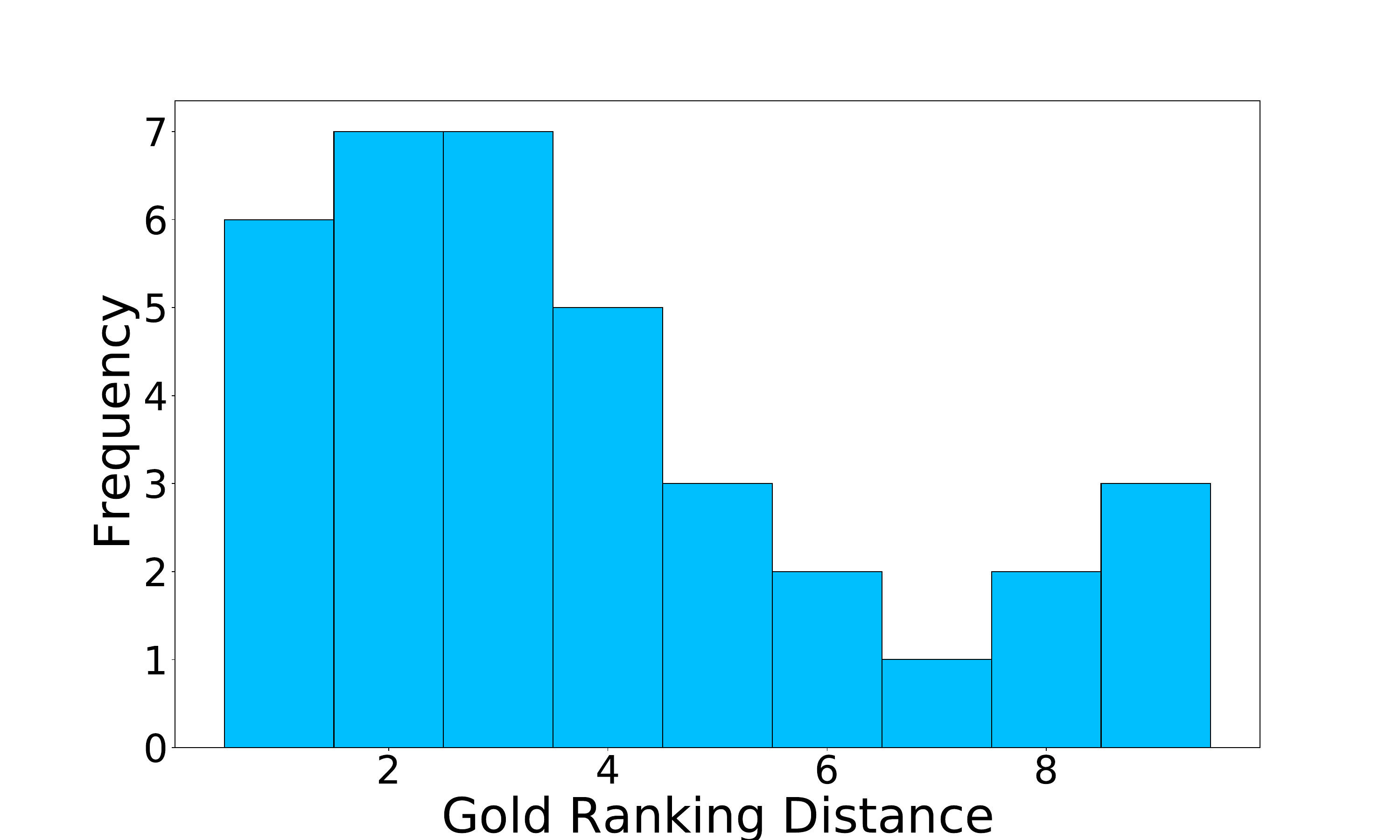}
        \caption{LLM GPT-4 - WMT'19 En-De}
    \end{subfigure}
    \caption{Histogram of distance between the items in the discordant pairs in the gold ranking for Neuron-wise MoS and LLM GPT-4 across three MT benchmarks. The discordant pairs of LLM-PP lie mostly around low gold ranking distances region (like Neuron-wise MoS), which should not ideally have a big negative impact for the NAS task.}
\label{fig:ktau-histogram-id-distance-wmt14ende-gpt-4}
\end{figure*}

\begin{figure*}[t!]
    \centering
    \begin{subfigure}[t]{0.32\textwidth}
        \centering
        \includegraphics[height=1.6in, width=2.0in]{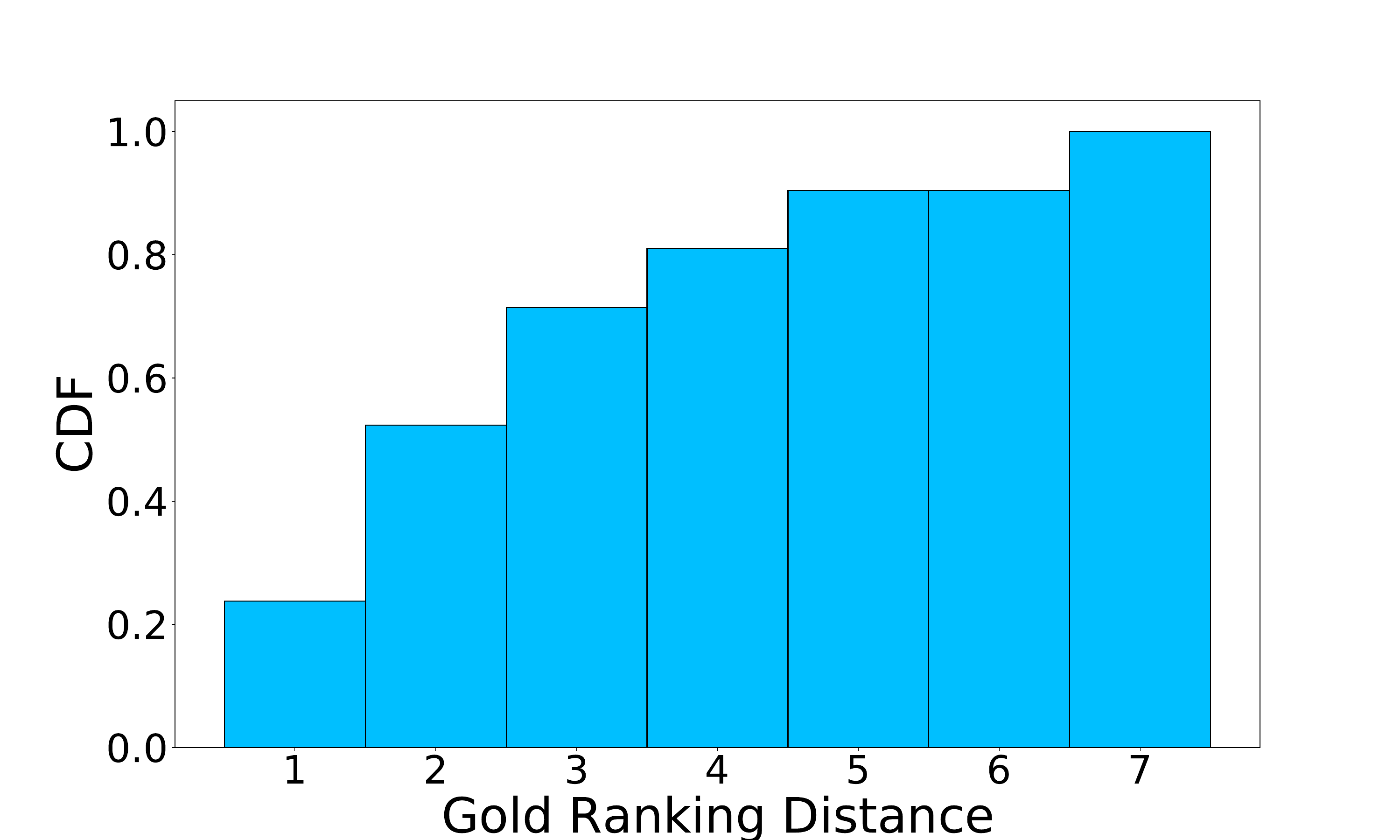}
        \caption{Neur. MoS - WMT'14 En-De}
    \end{subfigure}%
    ~ 
    \begin{subfigure}[t]{0.32\textwidth}
        \centering
        \includegraphics[height=1.6in, width=2.0in]{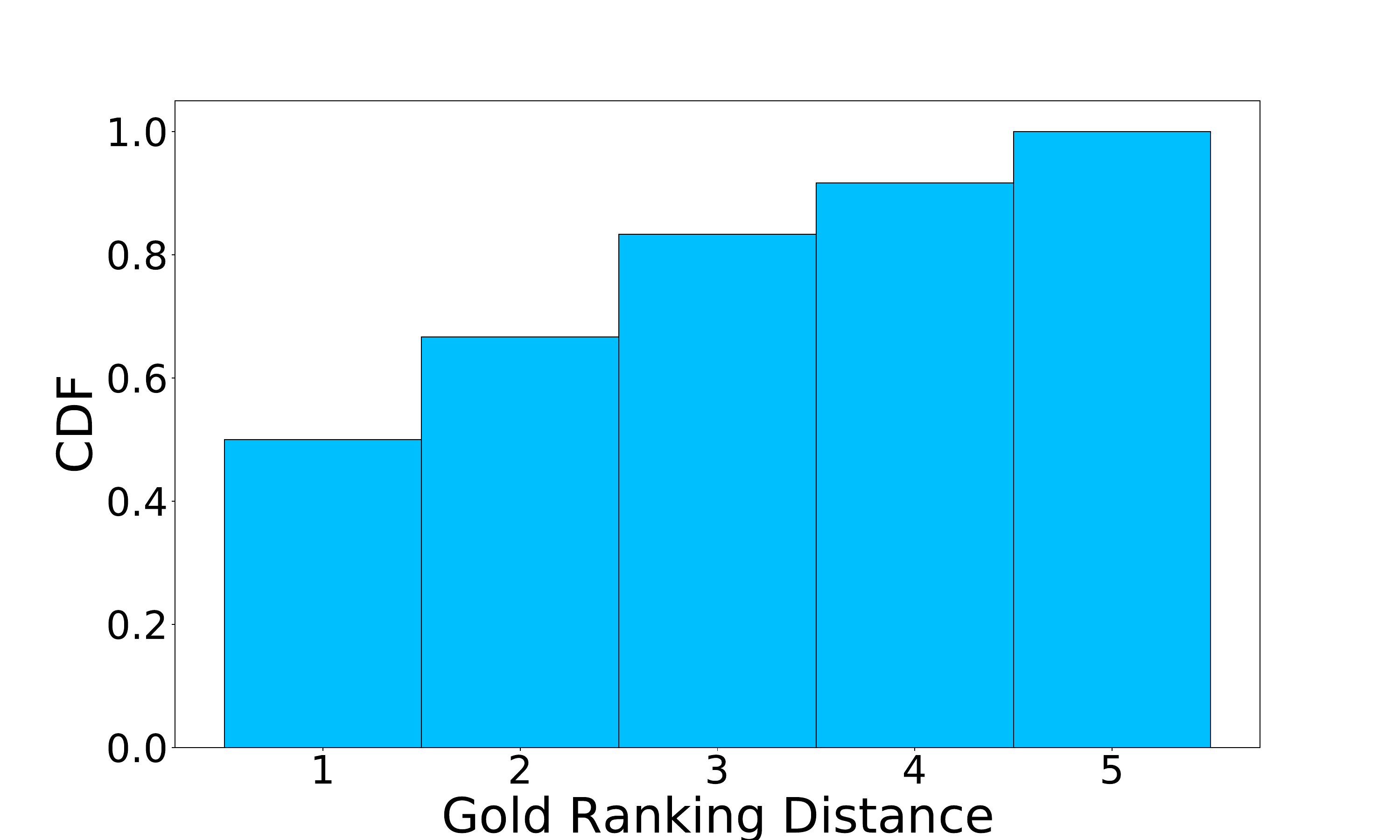}
        \caption{Neur. MoS - WMT'14 En-Fr}
    \end{subfigure}
    ~
    \begin{subfigure}[t]{0.32\textwidth}
        \centering
        \includegraphics[height=1.6in, width=2.0in]{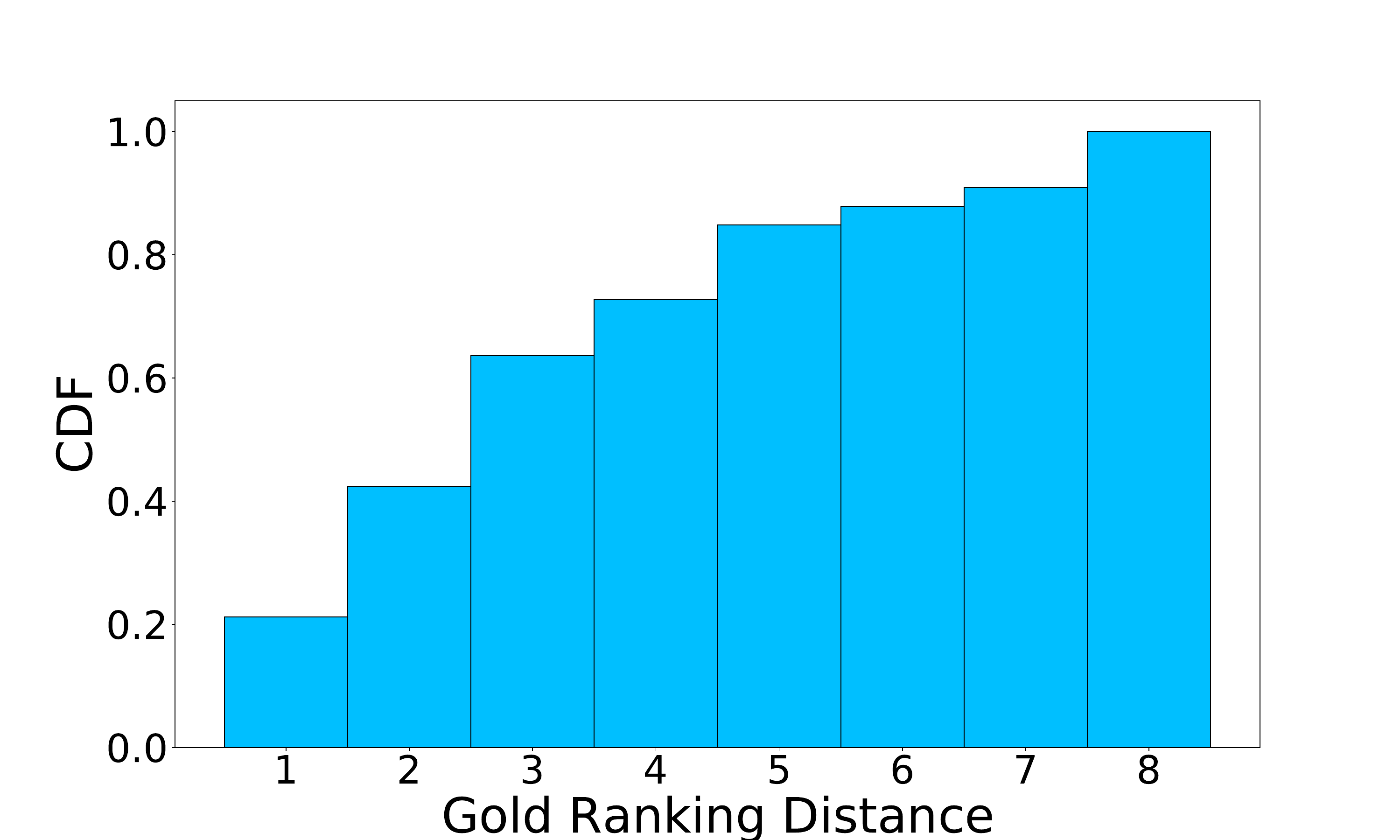}
        \caption{Neur. MoS - WMT'19 En-De}
    \end{subfigure}
    \begin{subfigure}[t]{0.32\textwidth}
        \centering
        \includegraphics[height=1.6in, width=2.0in]{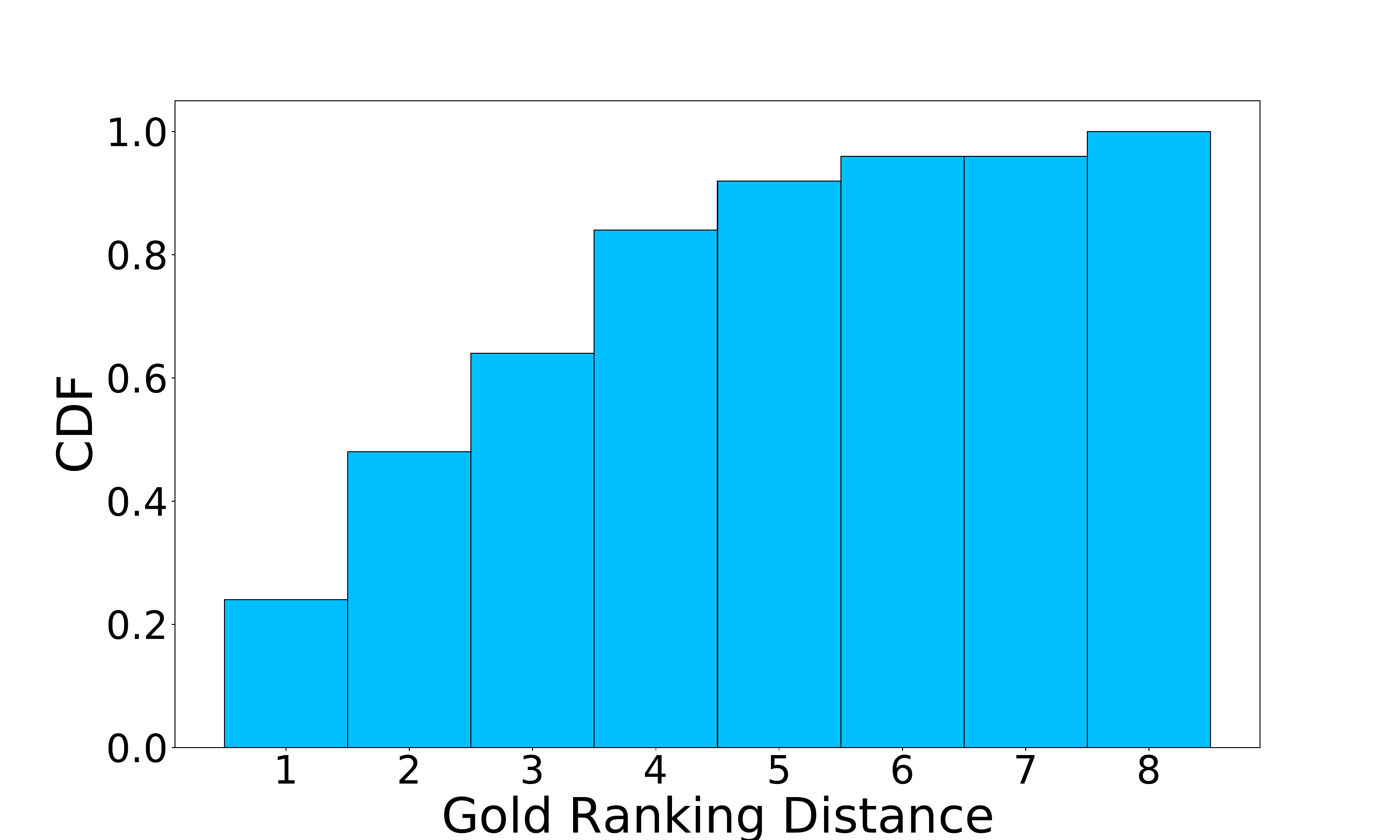}
        \caption{LLM GPT-4 - WMT'14 En-De}
    \end{subfigure}%
    ~ 
    \begin{subfigure}[t]{0.32\textwidth}
        \centering
        \includegraphics[height=1.6in, width=2.0in]{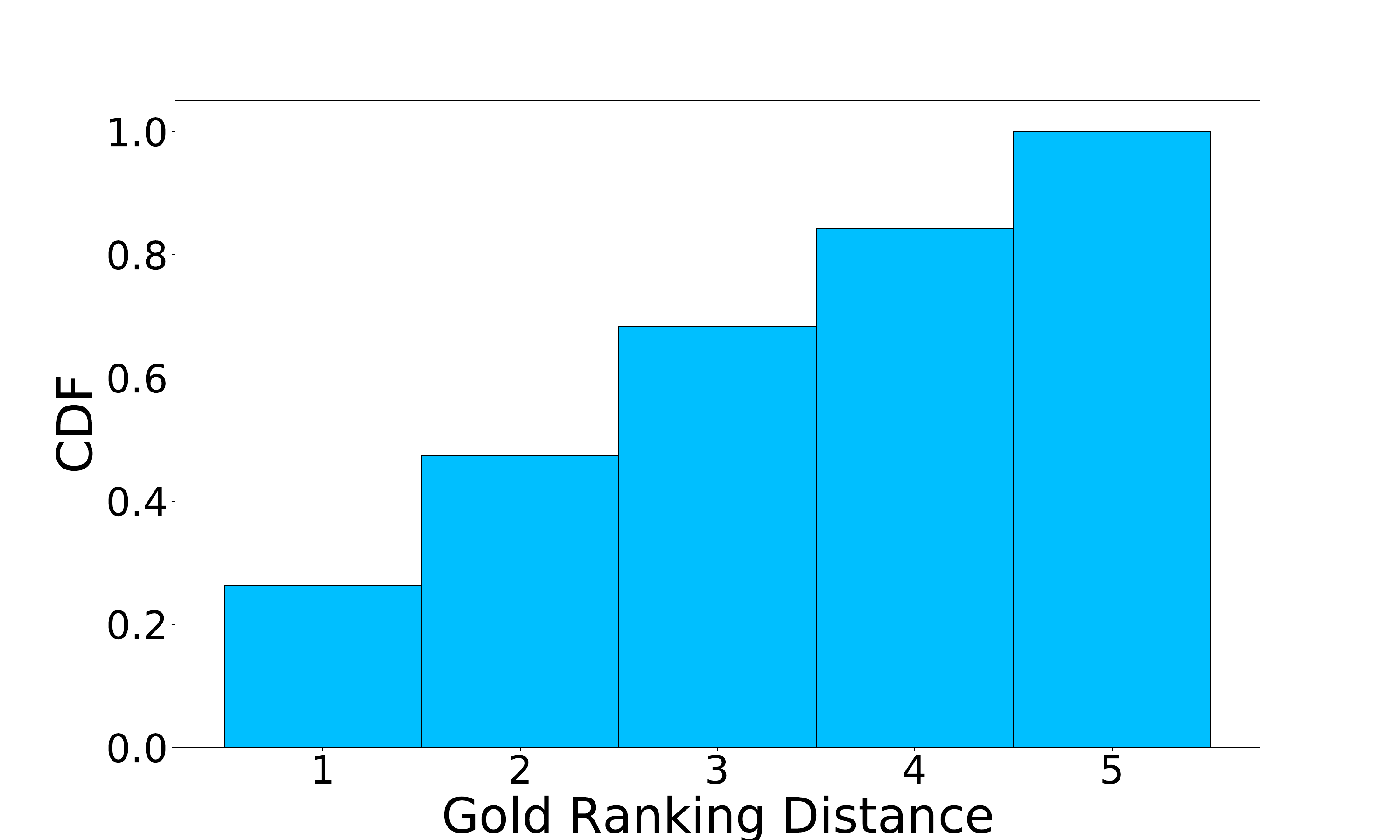}
        \caption{LLM GPT-4 - WMT'14 En-Fr}
    \end{subfigure}
    ~
    \begin{subfigure}[t]{0.32\textwidth}
        \centering
        \includegraphics[height=1.6in, width=2.0in]{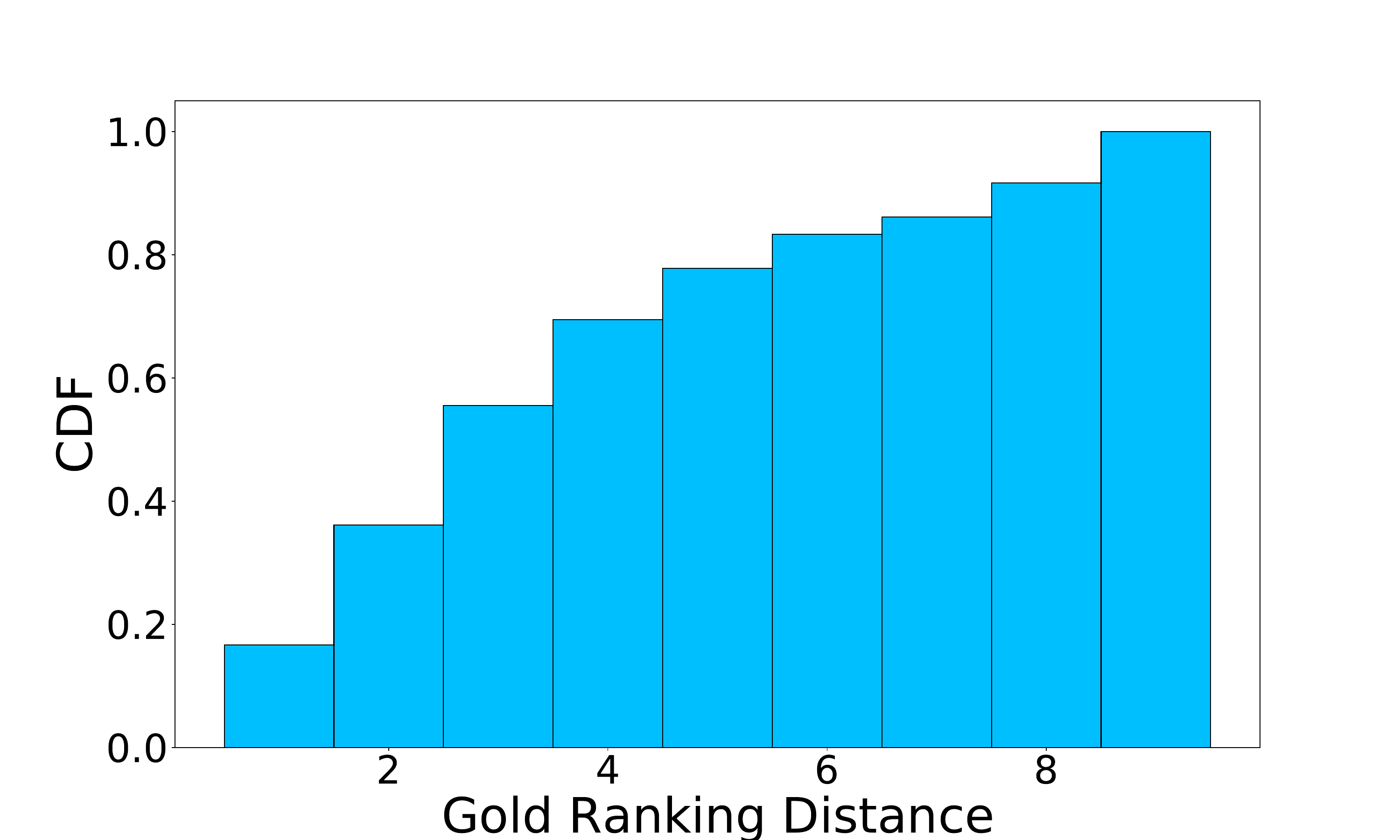}
        \caption{LLM GPT-4 - WMT'19 En-De}
    \end{subfigure}
    \caption{Cummulative distribution function of distance between the items in the discordant pairs in the gold ranking for Neuron-wise MoS and LLM GPT-4 across three MT benchmarks. The cummulative distribution function of gold ranking distances for discordant pairs for LLM-PP GPT-4 and Neuron-wise MoS are very similar.}
\label{fig:ktau-cdf-histogram-id-distance-wmt14ende-gpt-4}
\end{figure*}

We perform a fine-grained analysis of Kendall-Tau performance for Neuron-wise MoS and LLM-PP GPT-4. In figure~\ref{fig:ktau-histogram-id-distance-wmt14ende-gpt-4}, we plot the histogram of distance between the items in the discordant pairs in the gold ranking for Neuron-wise MoS and LLM GPT-4 across three MT benchmarks. The discordant pairs of LLM-PP lie mostly around low gold ranking distances region (like Neuron-wise MoS), which should not ideally have a big negative impact for the NAS task. In figure~\ref{fig:ktau-cdf-histogram-id-distance-wmt14ende-gpt-4}, we plot the corresponding cummulative distribution function (CDF). The CDF of gold ranking distances for discordant pairs for LLM-PP GPT-4 and Neuron-wise MoS are very similar.

\subsection{Machine Translation Details}

\subsubsection{Machine Translation - Dataset Statistics}
\label{sec:mt_dataset_statistics}

The statistics of the MT benchmarks is shown in Table~\ref{tab:dataset_statistics}.

\begin{table*}
\scriptsize
\begin{center}
\begin{tabular}{lllllll}
\toprule
\textbf{Dataset} & \textbf{Year} & \textbf{Source Lang} & \textbf{Target Lang} & \textbf{\#Train} & \textbf{\#Valid} & \textbf{\#Test} \\ \midrule
WMT & 2014 & English (en) & German (de) & 4.5M & 3000 & 3000 \\
WMT & 2019 & English (en) & German (de) & 43M & 2900 & 2900 \\
WMT & 2014 & English (en) & French (fr) & 35M & 26000 & 26000 \\
\bottomrule
\end{tabular}
\caption{Statistics - Machine translation benchmark.}
\label{tab:dataset_statistics}
\end{center}
\end{table*}

\subsubsection{Machine Translation - Training Details and Search Space}
\label{sec:mt_train_det_search_space}

Settings for training machine translation model include: $40K$ training steps, a cosine learning rate scheduler, Adam optimizer, and a warmup of learning rate from $10^{-7}$ to $10^{-3}$ with cosine annealing. The validation loss is used for model selection. The beam size is four with length penalty of 0.6. The search space ($\mathcal{A}$) is borrowed from HAT~\citep{hat}, which is also shown in Table~\ref{tab:design_space}.

\subsubsection{Architecture Encoding}
\label{sec:mt_arch_encoding}
Each machine translation architecture is encoded using a list of following 10 values:
\begin{enumerate}
    \item \textit{Encoder embedding dimension} corresponds to embedding dimension of the encoder.
    \item \textit{Encoder \#layers} corresponds to number of encoder layers.
    \item \textit{Average encoder FFN. intermediate dimension} corresponds to average of FFN intermediate dimension across encoder layers.
    \item \textit{Average encoder self attention heads} corresponds to average of number of self attention heads across encoder layers.
    \item \textit{Decoder embedding dimension} corresponds to embedding dimension of the decoder.
    \item \textit{Decoder \#Layers} corresponds to number of decoder layers.
    \item \textit{Average Decoder FFN. Intermediate Dimension} corresponds to average of FFN intermediate dimension across decoder layers.
    \item \textit{Average decoder self attention heads} corresponds to average of number of self attention heads across decoder layers.
    \item \textit{Average decoder cross attention heads} corresponds to average of number of cross attention heads across decoder layers.
    \item \textit{Average arbitrary encoder decoder attention} corresponds to average number of encoder layers attended by cross-attention heads in each decoder layer (-1 means only attend to the last layer, 1 means attend to the last two layers, 2 means attend to the last three layers).
\end{enumerate}

\begin{table}
\scriptsize
\begin{center}
\begin{tabular}{p{1.8in}p{0.8in}} \toprule
\textbf{Hyperparameter Attribute} & \textbf{Value choices} \\ \midrule
 Encoder-Embedding-Dim & \{512, 640\} \\
 Decoder-Embedding-Dim & \{512, 640\} \\
 \#Encoder-Layers & \{6\} \\
 \#Decoder-Layers & \{1, 2, 3, 4, 5, 6\}  \\
 Encoder-QKV-Dim & \{512\} \\
 Decoder-QKV-Dim & \{512\}  \\
 \#Encoder-Self-Attention-Heads (PL) & \{4, 8\}  \\
 \#Decoder-Self-Attention-Heads (PL) & \{4, 8\} \\
 \#Decoder-Cross-Attention-Heads (PL) & \{4, 8\}  \\
 \#Decoder-Arbitrary-Attention (PL) &\{-1, 1, 2\} \\ 
 Encoder-FFN-Intermediate-Embed-Dim (PL) & \{1024, 2048, 3072\} \\
 Decoder-FFN-Intermediate-Embed-Dim (PL) & \{1024, 2048, 3072\} \\ 
 \bottomrule
\end{tabular}
\caption{Search space ($\mathcal{A}$), borrowed from HAT~\citep{hat}. `PL' refers to hyperparameters that vary per layer.}
\label{tab:design_space}
\end{center}
\end{table}

\subsection{Search and Evaluation Setup - Details}
\label{sec:search_and_eval_details}
The hyperparameters of HS-NAS's search algorithm are taken from HAT: $\texttt{\mbox{num-iterations=30}}$, $\texttt{\mbox{population-size=125}}$, $\texttt{\mbox{num-parents=25}}$, $\texttt{\mbox{num-mutations=50}}$, $\texttt{\mbox{num-crossover=50}}$, and $\texttt{\mbox{mutate-prob=0.3}}$. We experiment with three $\texttt{\mbox{latency-constraint}}$s: $100ms$, $150ms$, and $200ms$. Once the search returns the best architecture, the final weights for this architecture is obtained by training the architecture from scratch to convergence using HAT's training settings (see~\ref{sec:mt_train_det_search_space}). The target hardware for search is NVIDIA V100 GPU with 32GB GPU RAM. The efficiency metric for search is  search hours, which accounts for the time taken to complete all the search iterations. We focus on the following architecture-specific efficiency metrics: (i) \textit{latency} - time taken in milliseconds to encode a sentence in source language and generate the translation in target language, (ii) \textit{GFLOPs} - gigaFLOPs taken for the feedforward propagation, and (iii) \textit{model size} - number of architecture-specific parameters in millions. Scripts to compute these metrics are taken from HAT's codebase~\footnote{\url{https://github.com/mit-han-lab/hardware-aware-transformers}} and we refer readers to the HAT paper for more details about how these metrics are computed.

\subsection{LLM-PP - Extended Results}
\label{sec:llm_pp_extended_results}

\subsubsection{LLM-PP vs. non-supernet baselines.} 
\label{sec:llmpp_vs_nonsupernet}

\begin{table}
\scriptsize
\begin{center}
\begin{tabular}{lp{0.6in}p{0.6in}p{0.6in}} \toprule
\textbf{Kendall-Tau} & \textbf{WMT'14 En-De} & \textbf{WMT'14 En-Fr} & \textbf{WMT'19 En-De} \\ \midrule
\# Params & 0.42 & 0.51 & 0.54 \\
\# FLOPs & 0.43	& 0.53 & 0.54 \\
grad-norm & -0.42 & -0.42 & -0.52 \\
snip & -0.42 & -0.27 & -0.3 \\
synflow & -0.31 & -0.47 & -0.49 \\
LLM-PP GPT-4 & \textbf{0.65} & \textbf{0.75} & \textbf{0.65} \\
\bottomrule
\end{tabular}
\caption{Kendall-Tau of LLM-PP GPT-4 vs. non-supernet baselines. LLM-PP beats non-supernet baselines as well.}
\label{tab:non-supernet-baselines}
\end{center}
\end{table}

{LLM-PP beats non-supernet baselines as well.} \change{We add comparison to five non-supernet baselines: \#Params, \#FLOPs, grad-norm, snip, and snyflow (see \citeauthor{colin2022adeeperlook} for details). From Table~\ref{tab:non-supernet-baselines}, it is clear that LLM-PP GPT-4 achieves a high Kendall Tau, outperforming all the non-supernet baselines. These results along with Table~\ref{tab:llm_pe_mae_tau_comparison} showcases the superior performance of LLM-PP across a wide range of baselines.}

\subsubsection{LLM-PP on recent datasets and low-resource/indigenous languages.} 
\label{sec:llmpp_recent_datasets}

\begin{table*}
\tiny
\begin{center}
\begin{tabular}{ccccccc} 
\toprule
\textbf{Dataset} & \multicolumn{2}{c}{\textbf{Bribri to Spanish}} & \multicolumn{2}{c}{\textbf{Chatino to Spanish}} & \multicolumn{2}{c}{\textbf{Spanish to Bribri}}  \\
\textbf{Performance Predictor} & \textbf{MAE } & \textbf{Kendall } & \textbf{MAE } & \textbf{Kendall } & \textbf{MAE } & \textbf{Kendall }
\\ \midrule
HAT & 0.28	& 0.15	& 1.55 &	\textbf{0.16} &	0.72	&0.02   \\
Layer-wise MoS & 0.33&	-0.13	&2.42&	-0.17	&0.63&	-0.14  \\
Neuron-wise MoS & 0.29&	-0.35&	2.94	&-0.06	&0.43&	0.09  \\ 
LLM-PP GPT-4 & \textbf{0.16} &	\textbf{0.29}	& \textbf{1.21}	&0.08	& \textbf{0.32}	& \textbf{0.20} \\ 
\bottomrule
\end{tabular}
\caption{MAE and Kendall-Tau between the performance predictor performance and the TFS performance, across two different seeds. LLM-PP works well for recent datasets and low-resource/indigenous languages.}
\label{tab:llm_pp_recent_datasets}
\end{center}
\end{table*}

{LLM-PP works well for recent datasets and low-resource/indigenous languages.} \change{Compared to SoTA performance predictors, LLM-PP GPT-4 works well for recent datasets (e.g., 2023 benchmark), low-resource/indigenous languages (e.g., Bribri, Chatino). From the recent shared task: ``AmericasNLP 2023 Shared Task on Machine Translation into Indigenous Languages''~\cite{ebrahimi-etal-2023-findings}, we take three machine translation benchmarks: Bribri to Spanish, Chatino to Spanish, and Spanish to Bribri. Compared to WMT 2014, WMT 2019 benchmarks, these three benchmarks are very recent (2023 year) and one of the languages in each translation direction is an low-resource/indigenous language (Bribri, Chatino). As shown in Table~\ref{tab:llm_pp_recent_datasets}, we compare LLM-PP GPT-4 against SoTA performance (BLEU) predictors on these benchmarks in terms of quality (MAE, Kendall-Tau). It is clear that LLM-PP achieves the SoTA MAE score across these benchmarks, which is consistent with the trends in WMT 2014, WMT 2019 benchmarks (as shown in Table~\ref{tab:llm_pe_mae_tau_comparison}). Impressively, on two of these benchmarks, LLM-PP also achieves the SoTA Kendall-Tau score. Put together, these results clearly showcase that LLM-PP generalizes well to recent datasets and low-resource languages.}

\subsubsection{LLM-PP for COMET metric.} 
\label{sec:llmpp_comet_metric}

\begin{table*}
\tiny
\begin{center}
\begin{tabular}{ccccc} 
\toprule
\textbf{Dataset} & \multicolumn{2}{c}{\textbf{Bribri to Spanish}} & \multicolumn{2}{c}{\textbf{Chatino to Spanish}}  \\
\textbf{Performance Predictor} & \textbf{MAE } & \textbf{Kendall } & \textbf{MAE } & \textbf{Kendall } 
\\ \midrule
HAT & 0.03	& 0.24	& 0.02	&  -0.15   \\
Layer-wise MoS & 0.02&	-0.05&	0.02&	0.26  \\
Neuron-wise MoS & 0.02&	0.32&	0.01&	0.34  \\ 
LLM-PP GPT-4 & \textbf{0.01} & \textbf{0.32}& \textbf{0.01} &	\textbf{0.54} \\ 
\bottomrule
\end{tabular}
\caption{MAE and Kendall-Tau between the performance predictor performance and the TFS performance for COMET metric, across two different seeds. LLM-PP generalizes well to uncommon evaluation metrics like COMET.}
\label{tab:llm_pp_comet}
\end{center}
\end{table*}

{LLM-PP generalizes well to uncommon evaluation metrics. 
\change{
We build performance predictors that predict the Crosslingual Optimized Metric for Evaluation of Translation (COMET)~\cite{rei-etal-2022-comet} (Unbabel/wmt22-comet-da), which is relatively newer than the BLEU metric. Consider the Table~\ref{tab:llm_pp_comet} (performance averaged across two seeds), on the Bribri to Spanish task and the Chatino to Spanish task, LLM-PP achieves the SoTA MAE and SoTA Kendall Tau performance compared to SoTA performance predictors. These results show that LLM-PP generalizes well to uncommon evaluation metrics like COMET. Note that we exclude Spanish to BriBri task, since COMET does not support Bribri.}

\subsubsection{LLM-PP for robust predictions.} 
\label{sec:llmpp_robust_preds}

{LLM-PP provides fairly robust performance predictions.} \change{We compute the predictions for 8500 randomly sampled architectures using LLM-PP GPT-4 three times and compute the standard deviation of the three predictions for each architecture. The mean of the standard deviation for 8500 architectures is very low: 0.21, 0.27, 0.27 BLEU for WMT’14 En-De, WMT’14 En-Fr, and WMT’19 En-De respectively. Thus, LLM-PP provides fairly robust performance predictions. For all our search experiments, we use a single estimate from LLM-PP.}

\subsection{HS-NAS - Expanded Algorithm}
\label{sec:hsnas_algo}

The expanded algorithm for HS-NAS can be found in Algorithm~\ref{algo:hsnasalgo}.

\begin{algorithm*}
\textbf{Input:} \\LLM-Distill-PP model: $\texttt{\mbox{llm-distill-pp}}$, \\ Weight-sharing supernet: $\texttt{\mbox{supernet}}$, \\ Latency predictor: $\texttt{\mbox{latency-predictor}}$, \\ \#Search iterations: $\texttt{\mbox{num-iterations}}$, \\ Population size: $\texttt{\mbox{population-size}}$, \\ \#Parents: $\texttt{\mbox{num-parents}}$, \\ \#Mutations: $\texttt{\mbox{num-mutations}}$, \\ \#Crossovers: $\texttt{\mbox{num-crossover}}$, \\ Mutate probability: $\texttt{\mbox{mutate-prob}}$, \\ Latency constraint: $\texttt{\mbox{latency-constraint}}$, \\ LLM-Distill-PP Start Iteration: $\texttt{\mbox{llm-start-iteration}}$, \\ LLM-Distill-PP End Iteration: $\texttt{\mbox{llm-end-iteration}}$  \\
\textbf{Output:} $\texttt{\mbox{best-architecture}}$ 
     \begin{algorithmic}[1]
     \State $popu \gets$ $\texttt{\mbox{population-size}}$ rand. samples from  search space {\textcolor{blue}{// create init. population}}
     \For{$iter \gets 1 \: \texttt{\mbox{to}} \: \texttt{\mbox{num-iterations}}$}
        \State {\textcolor{blue}{// gen. parents by picking top cand. arch.}}
        \If{{\textcolor{red}{ $\texttt{\mbox{llm-start-iteration}} < iter < \texttt{\mbox{llm-end-iteration}} $}} }
          \State {\textcolor{red}{$\texttt{\mbox{parents}} \gets$ top `$\texttt{\mbox{num-parents}}$' arch. from $popu$ by $\texttt{\mbox{llm-distill-pp}}$ }}
        \Else
          \State {\textcolor{red}{$\texttt{\mbox{parents}} \gets$ top `$\texttt{\mbox{num-parents}}$' arch. from $popu$ by $\texttt{\mbox{supernet}}$ }}
        \EndIf
        \State {\textcolor{blue}{// gen. cand. via mutation}}
        \State $\texttt{\mbox{mutate-popu}}$ = $\{ \}$  
        \For{$mi \gets 1 \: \texttt{\mbox{to}} \: \texttt{\mbox{num-mutations}}$}
            \State $\texttt{\mbox{gene}} \gets$ mutate a random eg from $popu$ with $\texttt{\mbox{mutate-prob}}$
            \If{$\texttt{\mbox{gene}}$ satisfies $\texttt{\mbox{latency-constraint}}$ via $\texttt{\mbox{latency-predictor}}$}
               \State $\texttt{\mbox{mutate-popu}} = \texttt{\mbox{mutate-popu}} \cup \texttt{\mbox{gene}}$
            \EndIf
        \EndFor
        \State {\textcolor{blue}{// gen. cand. via cross-over}}
        \State $\texttt{\mbox{crossover-popu}}$ = $\{ \}$ 
        \For{$ci \gets 1 \: \texttt{\mbox{to}} \: \texttt{\mbox{num-crossover}}$}
            \State $\texttt{\mbox{gene}} \gets$ crossover two random eg from $popu$
            \If{$\texttt{\mbox{gene}}$ satisfies $\texttt{\mbox{latency-constraint}}$ via $\texttt{\mbox{latency-predictor}}$}
               \State $\texttt{\mbox{crossover-popu}} = \texttt{\mbox{crossover-popu}} \cup \texttt{\mbox{gene}}$
            \EndIf
        \EndFor
        \State {\textcolor{blue}{// upd. population}}
        \State $popu = \texttt{\mbox{parents}} \cup \texttt{\mbox{mutate-popu}} \cup \texttt{\mbox{crossover-popu}}$ 
     \EndFor
     \State {{return top arch. from $popu$}}
\end{algorithmic}
\caption{Hybrid-Search algorithm for Neural Architecture Search (HS-NAS). Changes to HAT~\citep{hat}'s search algorithm are in {\textcolor{red}{red}} color.}
\label{algo:hsnasalgo}
\end{algorithm*}

\subsection{LLM-Distill-PP - Extended Results}
\label{sec:llm_distill_pp_extended_results}

\subsubsection{Performance predictor quality vs. prediction time.} 
\label{sec:llmdistillpp_perf_time}

\begin{table}
\scriptsize
\begin{center}
\begin{tabular}{lccc} \toprule
\textbf{Performance Predictor} & \textbf{MAE} & \textbf{Kendall-Tau} & \textbf{Prediction Time (s)} \\ \midrule
HAT & 1.14 & 0.71 & 10.5 \\
Layer-wise MoS & 1.05 & \textbf{0.81} &  13.9 \\
Neuron-wise MoS & 0.97 &  0.56 &  13.3 \\
LLM-PP GPT-4 &  0.28 &  0.65 &  11.9 \\
LLM-Distill-PP GPT-4  &  \textbf{0.22} &  {0.64} &  \textbf{0.01} \\
 \bottomrule
\end{tabular}
\caption{Performance predictor quality vs. prediction time.}
\label{tab:pp_eff_accuracy}
\end{center}
\end{table}

\change{Table~\ref{tab:pp_eff_accuracy} shows the efficiency (time taken to predict performance for 10 architectures) and accuracy (MAE, Kendall) for supernet-based PP (HAT, Layer-wise MoS, Neuron-wise MoS), LLM-PP (GPT-4), and LLM-Distill-PP (GPT-4). LLM-Distill-PP provides the best efficiency-accuracy tradeoff with on par accuracy as LLM-PP but significantly faster prediction time (0.01s vs. 11.9s).}

\subsubsection{Varying initialization seeds.} 
\label{sec:var_init_seeds}
{HS-NAS seems robust to initialization effects caused by different seeds, achieving largely similar numbers on all metrics.} This result is shown in Table~\ref{tab:nas_main_results_across_seeds}, where latency numbers change slightly while numbers for other metrics are almost the same.

\begin{table*}
\scriptsize
\begin{center}
\begin{tabular}{cccccc}
\toprule
\textbf{Seed} & \textbf{BLEU ($\uparrow$)} & \textbf{Latency (ms) ($\downarrow$)} & \textbf{GFLOPs ($\downarrow$)} &  \textbf{Model Size (M) ($\downarrow$)} &  \textbf{Search Hours ($\downarrow$)} \\ \midrule
$\mathbf{100ms}$ \\
1 & 40.7 & 104.1 & 2.54 & 63.8 & 3.14 \\
2 & 40.7 & 98.2 & 2.54 & 63.8 & 3.15 \\
3 & 40.7 & 101.2 & 2.58 & 63.8 & 3.16 \\ \midrule
$\mathbf{150ms}$ \\
1 & 41.5 & 160.4 & 3.35 & 74.3 & 3.89 \\
2 & 41.4 & 172.6 & 3.31 & 74.3 & 3.69 \\
3 & 41.5 & 158.5 & 3.35 & 74.3 & 3.84 \\
\bottomrule
\end{tabular}
\caption{Initialization effects of HS-NAS (GPT-4, HAT, 1, 15) on WMT'14 En-Fr for different latency constraints - Test BLEU, latency in milliseconds, GFLOPs, model size in millions, and search hours. HS-NAS seems robust to initialization effects, achieving similar numbers on all metrics of interest.}
\label{tab:nas_main_results_across_seeds}
\end{center}
\end{table*}

\subsubsection{Varying FLOPs constraints.} 
\label{sec:var_flop_constraints}
{HS-NAS performs similarly to HAT for different FLOPs constraints, with at least 16\% reduction in search hours, 1.2\% improvement in latency, same GFLOPs and same model size.} Table~\ref{tab:nas_main_results_across_flops_constraints} contains these superior results of HS-NAS across 2.5 and 3.0 GFLOPs constraints. These trends largely hold true across benchmarks as well, as shown in Table~\ref{tab:nas_main_results_across_flops_constraints_across_datasets}.

\begin{table*}
\scriptsize
\begin{center}
\begin{tabular}{cccccc}
\toprule
\textbf{Search} & \textbf{BLEU ($\uparrow$)} & \textbf{Latency (ms) ($\downarrow$)} & \textbf{GFLOPs ($\downarrow$)} &  \textbf{Model Size (M) ($\downarrow$)} &  \textbf{Search Hours ($\downarrow$)} \\ \midrule
$\textbf{2.5 GFLOPs}$ \\
HAT & \textbf{26.9} & 69.5 & \textbf{2.47} & \textbf{41.0} & 2.54 \\
HS-NAS & 26.7 & \textbf{68.6} & \textbf{2.47} & \textbf{41.0} & \textbf{2.13} \\
\midrule
$\textbf{3.0 GFLOPs}$ \\
HAT & 27.5 & 125.4 & \textbf{2.98} & \textbf{49.4} & 2.08 \\
HS-NAS & \textbf{27.6} & \textbf{123.9} & \textbf{2.98} & \textbf{49.4} & \textbf{1.51} \\
\bottomrule
\end{tabular}
\caption{HS-NAS (GPT-4, HAT, 1, 15) vs. HAT on WMT'14 En-De for different FLOPs constraints - Test BLEU, latency in milliseconds, GFLOPs, model size in millions, and search hours. HS-NAS (GPT-4, HAT, 1, 15) performs similarly to HAT, with at least 16\% reduction in search hours, 1.2\% improvement in latency, same GFLOPs and same model size.}
\label{tab:nas_main_results_across_flops_constraints}
\end{center}
\end{table*}

\begin{table*}
\scriptsize
\begin{center}
\begin{tabular}{cllccl}
\toprule
\textbf{Search} & \textbf{BLEU ($\uparrow$)} & \textbf{Latency (ms) ($\downarrow$)} & \textbf{GFLOPs ($\downarrow$)} &  \textbf{Model Size (M) ($\downarrow$)} &  \textbf{Search Hours ($\downarrow$)} \\ \midrule
$\textbf{WMT'14 En-De}$ \\
HAT & 27.5 & 125.4 & \textbf{2.98} & \textbf{49.4} & 2.08 \\
HS-NAS & \textbf{27.6 (+0.4\%)} & \textbf{123.9 (-1.2\%)} & \textbf{2.98} & \textbf{49.4} & \textbf{1.51 (-27.4\%)}  \\
\midrule
$\textbf{WMT'14 En-Fr}$ \\
HAT & 39.4 & \textbf{69.6} & \textbf{2.99} & \textbf{49.1} & 6.69 \\
HS-NAS & \textbf{39.8 (+1\%)} & {96.8 (+39.1\%)} & \textbf{3} & \textbf{49.1} & \textbf{4.2 (-37.2\%)} \\ \midrule
$\textbf{WMT'19 En-De}$ \\
HAT & 42.9 & {85.5} & \textbf{2.99} & \textbf{49.6} & 2.35 \\
HS-NAS & \textbf{43.1 (+0.5\%)} & \textbf{71.9 (+15.9\%)} & \textbf{2.99} & \textbf{49.6} & \textbf{2.03 (-13.6\%)} \\
\bottomrule
\end{tabular}
\caption{HS-NAS (GPT-4, HAT, 1, 15) vs. HAT across benchmarks for 3.0 GFLOPs constraint - Test BLEU, latency in milliseconds, GFLOPs, model size in millions, and search hours. HS-NAS (GPT-4, HAT, 1, 15) performs similarly or better than HAT, with at least 13\% reduction in search hours, at least 1.2\% improvement in latency (in most cases), same GFLOPs, and same model size.}
\label{tab:nas_main_results_across_flops_constraints_across_datasets}
\end{center}
\end{table*}

\subsubsection{Varying underlying supernet.} 
\label{sec:var_underlying_supernet}
{The dominance of HS-NAS seems consistent across the underlying supernet.} In the results so far, HAT is the supernet used by HS-NAS. In Table~\ref{tab:hsnas_results_with_fallback_to_neuron_moe}, we replace HAT with Neuron-wise MoS and show that HS-NAS performs similarly to Neuron-wise MoS, with at least 50\% reduction in search hours,  better or similar model size and GFLOPs. 

\begin{table*}
\scriptsize
\begin{center}
\begin{tabular}{llllll}
\toprule
\textbf{Search} & \textbf{BLEU ($\uparrow$)} & \textbf{Latency (ms) ($\downarrow$)} & \textbf{GFLOPs ($\downarrow$)} &  \textbf{Model Size (M) ($\downarrow$)} &  \textbf{Search Hours ($\downarrow$)} \\ \midrule
$\mathbf{100ms}$ \\
Neuron-wise MoS & \textbf{40.9} & \textbf{97.6} & \textbf{3.13} & \textbf{70.5} & 7.03 \\
HS-NAS (GPT-4, Neur., 1, 15) & \textbf{40.9} & {126.9 (+30\%)} & \textbf{3.13} & \textbf{70.5} & \textbf{3.36 (-52.2\%)} \\ \midrule
$\mathbf{150ms}$  \\
Neuron-wise MoS & \textbf{41.4} & 200.2 & 4.26 & 92.8 & 8.35 \\
HS-NAS (GPT-4, Neur., 1, 15) & {41.3 (-0.2\%)} & \textbf{162.2 (19.0\%)} & \textbf{4.22 (-0.9\%)} & \textbf{91.5 (1.4\%)} & \textbf{4.14 (-50.4\%)}  \\ \midrule
$\mathbf{200ms}$  \\
Neuron-wise MoS & 41.6 & \textbf{184.1} & \textbf{4.53} & \textbf{99.4} & 8.77 \\
HS-NAS (GPT-4, Neur., 1, 15) & \textbf{41.7 (+0.2\%)} & 191.2 (+3.9\%) & \textbf{4.53} & \textbf{99.4} & \textbf{4.22 (-51.8\%)} \\
\bottomrule
\end{tabular}
\caption{HS-NAS (GPT-4, Neuron-wise MoS, 1, 15) versus SoTA NAS on WMT'14 En-Fr for different latency constraints - Test BLEU, latency in milliseconds, GFLOPs, model size in millions, and search hours. HS-NAS is accompanied by four arguments: ($\texttt{\mbox{llm-distill-pp}}$, $\texttt{\mbox{supernet}}$, $\texttt{\mbox{llm-start-iteration}}$, $\texttt{\mbox{llm-end-iteration}}$ ). Across latency constraints, HS-NAS performs similarly or improves upon SoTA NAS, with at least 50\% reduction in search hours,  better or similar model size and GFLOPs.}
\label{tab:hsnas_results_with_fallback_to_neuron_moe}
\end{center}
\end{table*}

\subsubsection{Trivially constructed efficient adaptations of SoTA} 
\label{sec:efficient_adaptations}
Search hours can be trivially reduced in several ways: halving the total number of search iterations and/or using distilled SoTA predictor instead of using supernet predictor directly. As shown in Table~\ref{tab:hsnas_vs_hatloss_predictor_wmt14ende}, the former approach suffers from a big drop in BLEU performance (1.8\% for HAT ($\texttt{\mbox{num-iter.}}$=15)), while the latter approach suffers from a big increase in latency and GFLOPs (9.7\% and 32\% respectively for Distilled HAT ($\texttt{\mbox{num-iter.}}$=15)). On the other hand, HS-NAS dominates these adaptions in search hour reductions, while maintaining the performance of SoTA and not degrading on any footprint metric.

\begin{table*}
\scriptsize
\begin{center}
\begin{tabular}{llllll}
\toprule
\textbf{Search} & \textbf{BLEU ($\uparrow$)} & \textbf{Latency (ms) ($\downarrow$)} & \textbf{GFLOPs ($\downarrow$)} &  \textbf{Model Size (M) ($\downarrow$)} &  \textbf{Search Hours ($\downarrow$)} \\ \midrule
HAT ($\texttt{\mbox{num-iter.}}$=30) & \textbf{27.9} & 102.0 & 3.0 & 64.4 & 1.09   \\
HAT ($\texttt{\mbox{num-iter.}}$=15) & 27.4 (-1.8\%) & 107.6 (+5.5\%) & \textbf{2.96 (-1.3\%)} & \textbf{63.1 (-2\%)} & 0.65 (-40.4\%) \\
Distilled HAT ($\texttt{\mbox{num-iter.}}$=15) & {27.8 (-0.4\%)} & 111.9 (+9.7\%) & 3.97 (+32\%) & \textbf{63.1 (-2\%)} & 0.58 (-46.8\%) \\
HS-NAS (GPT-4, HAT, 1, 15) & \textbf{27.9} & \textbf{99.7 (-2.3\%)} & \textbf{2.96 (-1.3\%)} & \textbf{63.1 (-2\%)} & \textbf{0.56 (-48.6\%)} \\ 
\bottomrule
\end{tabular}
\caption{HS-NAS versus trivial efficient adaptations of SoTA with half of the original search iterations (original $\texttt{\mbox{num-iterations}}=30$): \textit{original SoTA}, \textit{distilled SoTA} on WMT'14 En-De for 100ms latency constraint - Test BLEU, latency in milliseconds, GFLOPs, model size in millions, and search hours. HS-NAS is accompanied by four arguments: ($\texttt{\mbox{llm-distill-pp}}$, $\texttt{\mbox{supernet}}$, $\texttt{\mbox{llm-start-iteration}}$, $\texttt{\mbox{llm-end-iteration}}$). Efficient adaptations of SoTA reduce search hours by at least 40\%, at the expense of either a big drop in BLEU performance (1.8\% for HAT ($\texttt{\mbox{num-iter.}}$=15) ) or big increase in latency and GFLOPs (9.7\% and 32\% respectively for Distilled HAT ($\texttt{\mbox{num-iter.}}$=15)). On the other hand, HS-NAS dominates these adaptions in search hour reductions, while maintaining the performance of SoTA and not degrading on any footprint metric.}
\label{tab:hsnas_vs_hatloss_predictor_wmt14ende}
\end{center}
\end{table*}

